\documentclass[5p,times]{elsarticle}

\usepackage[utf8]{inputenc}
\usepackage{multirow}
\usepackage{float}
\usepackage{subfig} 
\usepackage{amsmath,amsfonts}
\usepackage{mathtools}
\usepackage{bm}
\usepackage{multirow}
\usepackage{enumitem}
\usepackage[ruled,vlined,linesnumbered,noend]{algorithm2e}
\usepackage[colorlinks=true, allcolors=blue]{hyperref}
\usepackage[table,xcdraw]{xcolor}

\usepackage{combelow}

\emergencystretch=\maxdimen
\newlength{\commentWidth}
\newcommand{\atcp}[1]{\tcp*[r]{\makebox[\commentWidth]{#1\hfill}}}

\begin{document}

\begin{frontmatter}

\title{GETAE: \underline{G}raph Information \underline{E}nhanced Deep Neural Ne\underline{t}work Ensemble \underline{A}rchitectur\underline{E} for Fake News Detection}
\author[1]{Ciprian-Octavian~Truică\fnref{c2}}
\cortext[c1]{Corresponding author.}
\fntext[c2]{These authors contributed equally to this article.}
\ead{ciprian.truica@upb.ro}

\author[1,2]{Elena-Simona Apostol\corref{c1}\fnref{c2}}
\ead{elena.apostol@upb.ro}

\author[1]{Marius Marogel\fnref{c2}}
\ead{marius.marogel@stud.acs.upb.ro}

\author[3,4]{Adrian Paschke}
\ead{paschke@inf.fu-berlin.de}

\affiliation[1]{organization={Computer Science and Engineering Department, Faculty of Automatic Control and Computers, National University of Science and Technology Politehnica Bucharest},
    addressline={Splaiul Independentei 313}, 
    city={Bucharest},
    postcode={060042}, 
    country={Romania}
}

\affiliation[2]{organization={Academy of Romanian Scientists},
    addressline={Ilfov 3}, 
    city={Bucharest},
    postcode={050044},     
    country={Romania}
}

\affiliation[3]{organization={Department of Mathematics and Computer Science, Freie Universität Berlin},
    addressline={Arnimallee 14}, 
    city={Berlin},
    postcode={14195}, 
    country={Germany}
}

\affiliation[4]{organization={Fraunhofer Institute for Open Communication Systems},
    addressline={Kaiserin-Augusta-Allee 31}, 
    city={Berlin},
    postcode={10589 }, 
    country={Germany}
}

\begin{abstract}
In today's digital age, fake news has become a major problem that has serious consequences, ranging from social unrest to political upheaval.
To address this issue, new methods for detecting and mitigating fake news are required.
In this work, we propose to incorporate contextual and network-aware features into the detection process.
This involves analyzing not only the content of a news article but also the context in which it was shared and the network of users who shared it, i.e., the information diffusion.
Thus, we propose GETAE,  \underline{G}raph Information \underline{E}nhanced Deep Neural Ne\underline{t}work Ensemble \underline{A}rchitectur\underline{E} for Fake News Detection, a novel ensemble architecture that uses textual content together with the social interactions to improve fake news detection.
GETAE contains two Branches: the Text Branch and the Propagation Branch.
The Text Branch uses Word and Transformer Embeddings and a Deep Neural Network based on feed-forward and bidirectional Recurrent Neural Networks (\textsc{[Bi]RNN}) for learning novel contextual features and creating a novel Text Content Embedding.
The Propagation Branch considers the information propagation within the graph network and proposes a Deep Learning architecture that employs Node Embeddings to create novel Propagation Embedding.
GETAE Ensemble combines the two novel embeddings, i.e., Text Content Embedding and Propagation Embedding, to create a novel \textit{Propagation-Enhanced Content Embedding} which is afterward used for classification.
The experimental results obtained on two real-world publicly available datasets, i.e., Twitter15 and Twitter16, prove that using this approach improves fake news detection and outperforms state-of-the-art models.
\end{abstract}


\end{frontmatter}

\section{Introduction}

As mass media becomes more digitalized, new journalistic models for information dissemination have emerged. 
These models have significantly transformed how society consumes news and information.
In the race to stay ahead of competitors, journalists sometimes sacrifice the standards of traditional journalism, prioritizing speed and aiming to ``go viral'' by quickly generating views aiming for an increase in likes, comments, and shares.
This new approach focuses on catering to users' needs, behaviors, and interests.
While digital media offers many benefits, it also heightens the risk of misinformation~\citep{Mustafaraj2017,Ruths2019}, which can have harmful consequences for society by enabling the rapid spread of false information, such as fake news related to the Brexit referendum~\citep{Bastos2017}, the 2016 US presidential election~\citep{Bovet2019}, and COVID-19 vaccinations~\citep{Rzymski2021}.

The current research explores models for detecting fake news, which are trained using either traditional machine learning or deep learning techniques on large datasets of news articles.
New approaches also incorporate additional metadata~\cite{Zhang2020}, such as citation sources, verifiable authors, short paragraphs, third-person narration, and other factors that can enhance the accuracy of fake news detection models.
However, training on such data can have drawbacks, especially when the model encounters new data that does not adhere to the same writing conventions. 
For example, social media content is typically more personal and informal~\cite{Aimeur2023}. 
This is why integrating extra information is crucial when developing fake news detection systems for social media.
Various methods~\cite{Truica2023danes} exist for representing features in this context, which can be divided into news-content features (e.g., headlines, text, images, or linguistic/visual elements) and social context features (e.g., user, post, or network-based data).
A major shortcoming in the current literature is that few current methods for analyzing social media data incorporate information propagation for the detection task.

To address this major shortcoming in the current literature, we propose GETAE a novel \underline{G}raph Information \underline{E}nhanced Deep Neural Ne\underline{t}work Ensemble \underline{A}rchitectur\underline{E} for Fake News Detection.
Our approach utilizes social network information stored as a graph, creating node embeddings with advanced algorithms.
These embeddings, combined with textual content, are then fed into feed-forward or bidirectional recurrent neural networks.
This enables our model to leverage both contextual and network-aware data for fake news detection.

The main objective of this work is to design and implement the novel \underline{G}raph Information \underline{E}nhanced Deep Neural Ne\underline{t}work Ensemble \underline{A}rchitectur\underline{E} for Fake News Detection, i.e., GETAE.
GETAE uses both textual content together with the social interactions to improve fake news detection.
Our secondary objectives are four-fold:
\begin{itemize}
    \item[1)] Propose a novel Text Content Embedding, a new vector representation for textual that combines multiple complex textual features such as lexical (e.g., character and word level features) and syntactic (e.g., sentence-level features) to improve prediction.
    \item[2)] Propose a new Propagation Embedding designed to encode the information diffusion, i.e., the spread of information from a node to its followers.
    \item[3)] Propose an innovative Propagation-Enhanced Content Embedding that encapsulates textual content, contextual data, and propagation information to improve detection performance.
    \item[4)] Benchmark GETAE by training multiple models using cross-validation, ablation testing, and hyperparameter tuning on two real-world publicly available datasets, i.e., Twitter15 and Twitter16.
    \item[5)] Compare the best GETAE models with current state-of-the-art models.
\end{itemize}

The main research questions we want to address with this work are:
\begin{enumerate}
  \item[(\textit{Q1})] How much dataset-reliant is the choice of models and algorithms for Fake News Detection?
  \item[(\textit{Q2})] What is the impact of word embeddings that combine multiple complex textual features such as lexical (e.g., character and word level features) and syntactic (e.g., sentence-level features), on the performance of Fake News Detection systems?
  \item[(\textit{Q3})] What is the impact of employing node embeddings that encode the information diffusion through social networks on the effectiveness of Fake News Detection methods?
  \item[(\textit{Q4})] What is the impact of embeddings that consider both the information diffusion and the textual content and context on the performance of Fake News Detection systems?
  \item[(\textit{Q5})] How are these models performing in comparison with state-of-the-art Fake News Detection models?
\end{enumerate}

This article is structured as follows.
In Section~\ref{sec:sota}, we briefly present the current state-of-the-art.
Section~\ref{sec:methodology} presents the methodology for preprocessing the data and designing the GETAE Ensemble architecture.
In Section~\ref{sec:results}, we present ablation, cross-validation, and hyperparameter tuning for GETAE as well as a comparison with the state-of-the-art models.
Section~\ref{sec:discussion} discusses our findings and the limitations of our architecture.
Finally, Section~\ref{sec:conclusions} concludes this work and hints at future research directions.

\section{Related Work}\label{sec:sota}

Textual data representation is a challenging task when it comes to generating linguistic features. 
Shu et al.~\cite{Shu2017} discuss the use of linguistic and semantic features such as n-grams, total word count, and punctuation that can provide meaningful representations for fake news detection.
Ahmad et al.~\cite{Ahmad2020} applied ensemble learning methods like Random Forests, Boosting, Bagging, and Voting Classifiers on the LIWC (Linguistic Inquiry and Word Count) dataset~\cite{Pennebaker2015}, achieving higher accuracy than traditional models across four datasets.
N-gram models for feature extraction also yielded strong results~\cite{Khan2021,Wynne2019}. 
These approaches use classical machine learning algorithms with TF-IDF (Term Frequency-Inverse Document Frequency) for n-gram extraction to train models.

The current literature on detecting harmful content primarily focuses on techniques such as word embeddings~\cite{Ilie2021}, character-level embeddings~\cite{Khan2021}, transformer embeddings~\cite{Petrescu2023, Truica2022Misrobaerta}, sentence transformers~\cite{Truica2022checktaht2022}, and document embeddings~\cite{Truica2023}.
Various deep learning architectures are used as classifiers for harmful content detection, with most state-of-the-art models relying on LSTMs (Long Short-Term Memory networks), GRUs (Gated Recurrent Units), and/or CNNs (Convolutional Neural Networks) as foundational components, as seen in studies such as \cite{Ilie2021, Truica2022Misrobaerta}.

Other effective linguistic approaches for extracting features for fake news detection include using contextual inputs such as the headline, first two sentences of a news article, and news content to generate representations~\cite{Borges2019}.
These features used for training models that employ BiLSTM (Bidirectional Long Short-Term Memory) and BiGRU (Bidirectional Gated Recurrent Unit), manage to capture long-term dependencies in sequences.
Other approaches use Hierarchical Recursive Neural Networks and transformer-based architecture to construct a linguistic tree of news content using an encoder-decoder model~\cite{Zhou2023}.
This linguistic tree is then processed by a series of BiGRU layers within the network.

Graph Neural Networks (GNN) use node-oriented data to capture and encode information from networks, unlike classical deep learning models that operate on sequential data.
Understanding local and global features in graphs can help solve a great number of problems in areas like biological or social networks.
Propagation-oriented methods use the networks to understand how features change when information is spread into a network.

\begin{figure*}
    \centering
    \includegraphics[width=1\linewidth]{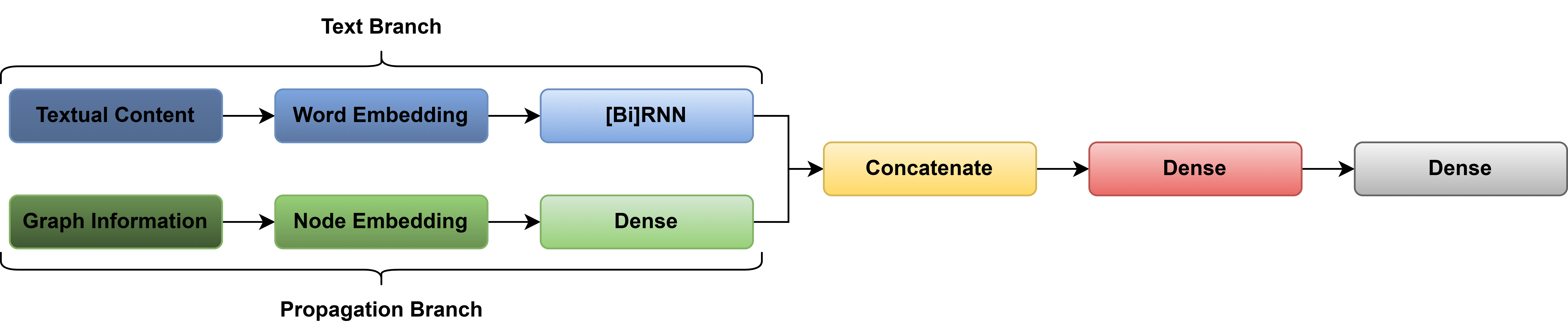}
    \caption{GETAE Architecture}
    \label{fig:enter-label}
\end{figure*}

Lu et al.~\cite{Lu2020GCAN} propose the graph-aware co-attention network, GCAN, a model that takes the user features in a propagation system and uses a GCN (Graph Convolutional Network) to learn representations for rumor detection.
Using a retweet order sorted based on the retweet time, they create GCN, GRU, and CNN embeddings for the tweet and social encodings, which are then passed through a dual co-attention mechanism.
This approach uses features from patterns in response time, source tweet content, and user characteristics on Twitter15 and Twitter16 datasets.

Huang et al.~\cite{Huang2019} use a deep structure learning method for rumor detection on the same Twitter-based datasets.
The authors create encoders for user data and propagation data with layers of GCN and RNN (Recursive Neural Networks), which are fed into an integrator module that uses both encoders to create representations of both user and propagation-tree data.
On rumor detection using semantic and propagation information, Ke et al.~\cite{Ke2020} propose KZWANG, a model that uses Multihead Attention mechanisms to generate a better representation for microblogs and GCN layers to fuse microblog features with propagation encoding.
This approach was tested on three datasets, surpassing other state-of-the-art models in accuracy and F1-score for multi-class classification. 

Topological information of a graph can be stored in a vector space, in the same way word embeddings are represented.
High-dimensional vector representations for nodes, edges, and whole graphs are noted as node, edge, or graph embeddings.
Goyal et al.~\cite{Goyal2018} and Cai et al.~\cite{Cai2018} use graph embeddings that encode node, edge, and whole graph information for solving different problems.
Xie et al.~\cite{Xie2020} use the GraphSage model for node embeddings, the Multihead Attention mechanism for creating user-based representations, and CNNs for content representations.
These layers are then concatenated for a binary classifier which gets good results on BuzzFeed and PolitiFact datasets, compared to other embedding techniques.
Imaduwage et al.~\cite{Imaduwage2022} propose a method for combining local and global graph-level embeddings. 
Each propagation tree contributes only partially with half priority with its local GraphSage embeddings to a global embedding mechanism.
These are fed to an autoencoder-based network that generates whole propagation-tree embeddings used then in a classification model.

When dealing with information diffusion on networks, the current literature proposes multiple solutions that 
analyze the spread of online content~\cite{Petrescu2019, Petrescu2024EDSA,Truica2021c} for both detection and network immunization tasks, such as preemptive~\cite{Petrescu2021}, community-based~\cite{Apostol2024CONTAIN}, or tree-based~\cite{Truica2023MCWDST} strategies.
Furthermore, the current literature also presents full solutions for real-time Fake News Detection. 
\textsc{StopHC}~\cite{Truica2024} is a full solution that builds on top of the current literature and proposes a novel deep-learning architecture for harmful content detection and mitigation.
\textsc{ContCommRTD}~\cite{Apostol2024ContCommRTD} is an architecture that detects misinformation on social media posts during real-time disaster reporting.
\citet{Petrescu2024} proposes a mixture of models to deal with harmful content detection.

\section{Methodology}\label{sec:methodology}

\subsection{Feature Representation}

\subsubsection{Text Preprocessing and Word Embeddings}

To preserve semantic relations while removing any elements that do not contribute to extracting the context (e.g., hyperlinks, utf8 characters, etc.), we process the textual data using a preprocessing pipeline.
We apply this preprocessing pipeline to reduce the vocabulary size~\cite{Truica2016a} before training while increasing the model's performance during inference~\cite{Truica2016Cats}.
By standardizing the textual data, we allow the model to focus on the core content and generalize better when unseen data is used for inference~\cite{Truica2017a}.
The preprocessing pipeline involves applying the following steps on the textual data~\cite{Truica2016a,Truica2017,Truica2021}: 
\textit{(1)} remove URL;
\textit{(2)} remove punctuation;
\textit{(3)} remove stopwords (only for the Word2Vec model); and
\textit{(4)} assign each token a unique integer identifier (id) starting with id $1$.


We reconstruct the textual data using the tokens' unique identifiers and obtain a 2-dimensional document-to-token matrix $D\in \mathbb{N}^{n \times k}$, where $n$ is the number of documents in the corpus and $k$ is the maximum size of a document in the corpus.
We use left zero padding for documents that are smaller than the maximum size $k$, i.e., we add $0$ on the left side of the document to achieve the maximum length $k$.

Using the tokenized text, we create 3 word embeddings: 
\begin{itemize}
    \item[\textit{(1)}] \textsc{Word2Vec} \textsc{Skip-Gram}~\cite{Mikolov2013}, a model that takes as input the word $w$ and tries to detect the context (the words on the left and right sides of $w$);
    \item[\textit{(2)}] BERT~\cite{Devlin2019} is a deep bidirectional transformer architecture that employs transfer learning to generalize language understanding;
    \item[\textit{(3)}] \textsc{BERTweet}~\cite{Nguyen2020} is a pre-trained language model for Twitter data that uses BERT's architecture and the pre-training procedure of RoBERTa.
\end{itemize}

We train our own \textsc{Word2Vec} embeddings and we use the pre-trained BERT and \textsc{BERTweet} transformer embeddings. 
Using these embeddings, we manage to better capture the relationships between tokens and encode local and global contexts~\cite{Truica2021}. 

We create a 2-dimensional matrix $W \in \mathbb{R}^{m \times s}$ that stores for each unique token its word embedding, where $m$ is the size of the vocabulary (i.e., the number of unique tokens in the corpus plus the padding element) and $s$ is the size of the word embedding (i.e., the number of the embedding components).
The word embedding for the padding element is a vector that contains only the value $0$.

\subsubsection{Graph Representation and Node Embeddings}

To construct a graph from a social network dataset, we use the users as nodes and the posting and user interaction with a post (i.e., re-sharing, linking, commenting, etc.) as edges.
Using this graph, we extract the information diffusion propagation graph $G=(V, E)$ of each node $u \in V$ to construct node embedding.
The diffusion is given by the edges $(u, v) \in E$ that connect 2 nodes $u, v \in V$.
For building node embeddings, we use \textsc{Node2Vec}~\cite{Grover2016} and \textsc{DeepWalk}~\cite{Perozzi2014}.
Both methods aim to capture the structural properties of the network and represent them in a low-dimensional space. 
\textsc{Node2Vec} is an extension of \textsc{DeepWalk} that uses a biased random walk to sample nodes in the network, which allows for the capture of both local and global network structures.

Specifically, \textsc{Node2Vec} uses a bias factor $\alpha_{p, q}$, and two parameters $p$ (return parameter) and $q$ (in-out parameter).
Equation~\eqref{eq:biasn2v} presents the calculation of the bias coefficient $\alpha_{p, q}$ when traversing a graph with a second-order random walk for the current node $u \in V$ by traversing edge $(u, v) \in E$, where $d_{u, v} \in \{ 0, 1, 2\}$ is the length of path of the edge $(u, v) \in E$. 

\begin{equation}\label{eq:biasn2v}
     \alpha_{p, q} = \begin{cases}
 & 1/p \text{, if $d_{u,v} = 0$} \\
 & 1   \text{, if $d_{u,v} = 1$} \\
 & 1/q \text{, if $d_{u,v} = 2$}  
    \end{cases}    
\end{equation}

Thus, the control of parameters $(p, q)$ influences the sampling decision of nodes that are more likely to produce similar embeddings.
The \textsc{Node2Vec} model uses a Skip-Gram with Negative Sampling (SGNS) approach to generate embeddings that are trained to maximize the likelihood $\mathcal{L}$ of predicting the neighborhood $p(N_s(u) \mid f(u)$ of a node given its embedding (Equation~\eqref{eq:n2vobj}), where $N_s(u)$ is the set of nodes sampled with strategy $s$ in the neighborhood of $u \in V$ and $f(u)$ is the mapping function from nodes to representations.

\begin{equation}\label{eq:n2vobj}
\mathcal{L} = \sum_{u \in V} \log p(N_s(u) \mid f(u)),
\end{equation}

\textsc{DeepWalk} uses a random walk to generate sequences of nodes and then applies the Skip-Gram model to learn node embeddings that capture the co-occurrence statistics of the nodes in the sequences.
\textsc{DeepWalk} is based on the idea that nodes that appear in similar contexts in the network are likely to have similar roles and functions, and therefore can be represented by similar embeddings.
Equation~\eqref{eq:dwobj} presents the objective function $\mathcal{L}$ that we try to maximize when creating embeddings with \textsc{DeepWalk}, where $N(u)$ is the set of nodes within the $k$-length random walk of node $u \in V$, and $f(u)$ and $f(v)$ are the embeddings for nodes $u, v \in V$.
In this case, the objective function $\mathcal{L}$ uses the \textit{softmax} function on the dot product of $f(u)$ and $f(v)$ to normalize the output in the interval $[0, 1]$.

\begin{equation}\label{eq:dwobj}
\mathcal{L} = \sum_{u \in V} \sum_{v \in N(u)} \log softmax(f(u) \cdot f(v)),
\end{equation} 

Both objective functions are optimized using stochastic gradient descent (SGD) with negative sampling. 

For each node embedding model, we create a 2-dimensional matrix $N \in \mathbb{R}^{|V| \times s}$ that stores the embedding vector of each node, where $|V|$ is the number of nodes in the graph and $s$ is the size of the node embedding (i.e., the number of the embedding components).

\subsection{GETAE Architecture}

\subsubsection{Text Branch}\label{ssec:text_branch}

GETAE's Text Branch contains 3 layers:
1) the input layer (i.e., Textual Content),
2) the Word Embedding layer, and
3) a hidden layer (i.e., \textsc{[Bi]RNN}).
The Textual Content layer requires 2 inputs:
1) the document-to-token matrix $D \in \mathbb{N}^{n \times k}$ and 
2) the word embedding matrix $W \in \mathbb{R}^{m \times s}$.
The Word Embedding layer is used to pair document tokens from $D$ to their corresponding word embeddings from $W$.
The hidden \textsc{[Bi]RNN} layer employs either Unidirectional Recurrent Neural Network (RNN) units or Bidirectional Recurrent Network (\textsc{BiRNN}) units.
In GETAE's implementation, the RNN is replaced by either Recurrent Neural Networks (RNN) units, Long Short-Term Memory (LSTM) units, or Gated Recurrent Units (GRU) for ablation testing.
We use the \textsc{[Bi]RNN} notation when discussing this layer, regardless if we use the RNN or the \textsc{BiRNN} layer.
Thus, when using an RNN hidden layer, the actual layer units employed by GETAE are either standard RNN, LSTM, or GRU.
When using a \textsc{BiRNN} hidden layer, the layer units are either standard \textsc{BiRNN}, \textsc{BiLSTM}, or \textsc{BiGRU}.
During ablation testing, we interchange unidirectional with bidirectional RNNs.
Thus, we aim to determine the importance of preserving the information from inputs when the information passes through the hidden state in a feed-forward manner (the unidirectional case) versus when the information is passed in both directions using a backward and forward approach (the bidirectional case).
The output of the \textsc{[Bi]RNN} hidden layer is a new word representation, i.e., Text Content Embedding.
The novel Text Content Embeddings combines multiple complex textual features such as lexical (e.g., character and word level features) and syntactic (e.g., sentence level features) used to improve prediction.

As we employ multiple units for the \textsc{[Bi]RNN} hidden layer, the Text Content Embeddings manage to create different types of embeddings that also consider the input Word Embedding.
Regardless of the word embedding used as input, we create 6 subtypes of Text Embeddings using \textsc{[Bi]RNN}: 
\begin{itemize}
    \item[(1)] Text Content Embedding using RNN: the word embedding passes through an RNN layer to obtain the final text embedding.
    \item[(2)] Text Content Embedding using BiRNN: the word embedding passes through a \textsc{BiRNN} layer to obtain the final text embedding.
    \item[(3)] Text Content Embedding using GRU: the word embedding passes through a GRU layer to obtain the final text embedding.
    \item[(4)] Text Content Embedding using BiGRU: the word embedding passes through a \textsc{BiGRU} layer to obtain the final text embedding.
    \item[(5)] Text Content Embedding using LSTM: the word embedding passes through an LSTM layer to obtain the final text embedding.
    \item[(6)] Text Content Embedding using BiLSTM: the word embedding passes through a \textsc{BiLSTM} layer to obtain the final text embedding. 
\end{itemize}

\subsubsection{Propagation Branch}\label{ssec:propagation_branch}

GETAE's Propagation Branch contains 3 layers:
1) the input layer (i.e., Graph Information),
2) Node Embedding layer, and
3) a hidden layer (i.e., Dense).
The Graph Information layer requires 2 inputs: 
1) list of nodes $V$ and 
2) the node embedding matrix $N \in \mathbb{R}^{|V| \times s}$.
The Node Embedding layer is used to pair nodes from $V$ to their corresponding node embeddings from $N$.
A hidden layer uses the pairs to obtain a novel Propagation Embedding.
This hidden layer contains Dense units that employ the \textit{ReLU} activation function to create a new Propagation Embedding.
The novel Propagation Embedding is designed to encode the information diffusion in order to encode the spread of information from a node to its followers. 

\subsubsection{Ensemble}\label{ssec:ensamble}

GETAE's Ensemble concatenates the output of the Text and Propagation Branches into one tensor, which is passed to a Dense layer used for classification.
During the concatenation, we create a new \textit{Propagation-Enhanced Content Embedding} that takes into account both the textual content and the information propagation within the social media graph, i.e., it combines the Text Content Embedding with the Propagation Embeddings.
This novel \textit{Propagation-Enhanced Content Embedding} is then used by a Dense layer that extracts hidden contextual features and creates a new vector representation by employing the \textit{ReLU} activation function.
The new vector representation is passed through the final Dense layer for classification which uses the \textit{softmax} activation function to determine the veracity of a post.

\subsubsection{Algorithm}

Algortihm~\ref{alg:getae} presents the pseudo-code for GETAE.
As input, GETAE takes the dataset $X=\{ <t_i, v_i, c_i> | i=\overline{1, n} \}$, the graph $G = (V, E)$, the configuration for the Text Branch given as a pair $<bidirection, reccurent>$, the word embedding model for the Text Branch $we\_model$, and the node embedding model for the Propagation Branch $ne\_model$.
The dataset $X$ contains for each observation $i=\overline{1, n}$ ($n=|X|$) a tuple $<t_i, v_i, c_i> $, where $t_i$ is the textual content, $v_i$ is the author of the content, i.e., node, and $c_i$ is the class.
The graph $G$ contains the list of nodes $V$ and the list of edges $E$ used to create the propagation graph of each node.
The configuration for the Text Branch determines the types of units used by the \textsc{[Bi]RNN} and if it is unidirectional or bidirectional as follows:
1) $bidirection$ is a boolean that indicates the use (True) or not (False) of a Bidirectional RNN, and
2) $reccurent$ is a Layer object that indicates what units are used, i.e., standard RNN, LSTM, or GRU.
The $we\_model$ determines which word embedding model is used, i.e., \textsc{Word2Vec}, BERT, or \textsc{BERTweet}.
Finally, $ne\_model$ determines which node embedding model is used, i.e., \textsc{Node2Vec} or \textsc{DeepWalk}.

In Lines~\ref{line01}-\ref{line07}, we initialize and populate the $T$, $V$, and $C$ lists where we store the text $t_i$, the nodes $v_i$, and the class $c_i$ for each entry in the dataset $X$.
After iterating through each record in $X$, the lists will look as follows: 
$T=\{t_i| i=\overline{1, n}\}$, 
$V=\{v_i | i=\overline{1, n}\}$, and 
$C=\{c_i| i=\overline{1, n}\}$. 
The list $T$ is used to obtain the document-to-token matrix $D$ (Line~\ref{line08}) and the word embedding matrix $W$ (Line~\ref{line09}).
The function $getWordEmbeding(T, we\_model)$ uses the $we\_model$ parameter to determine which word embedding model to use. 
The list $V$ is used to obtain the node embedding matrix $N$ (Line~\ref{line10}).
The function $getNodeEmbeding(G, V, ne\_model)$ uses the Graph and the $ne\_model$ parameter to determine which node embedding model to use. 
We note that the architecture is not limited to using only the presented word and node embeddings, as this is a modular architecture, any word and node embedding can be used.

The Text Branch is configured using the configuration pair $<bidirection, reccurent>$ (Lines~\ref{line11}-\ref{line27}), as explained in Subsection~\ref{ssec:text_branch}.
The input layer (Line~\ref{line11}) receives the document-to-token matrix $D$ and the word embedding matrix $W$.
This layer together with the Word Embedding layer is added to the Text Branch (Lines~\ref{line12}-\ref{line13}).
The \textsc{[Bi]RNN} layer is constructed in Lines~\ref{line14} to~\ref{line27}
Depending on parameter $bidirection$, the architecture uses either a feed-forward or a bidirectional RNN layer.
The type of units used by GETAE's RNN layer is selected based on the $reccurent$ parameter from the following types: standard RNN, LSTM, or GRU.

The Propagation Branch is created similarly to the Text Branch.
Lines~\ref{line28}-\ref{line31} present the layers of the Propagation Branch, as explained in Subsection~\ref{ssec:propagation_branch}.
The input layer (Line~\ref{line29}) receives the nodes list $V$ and the node embedding matrix $N$.
This layer together with the Node Embedding layer is added to the Propagation Branch (Lines~\ref{line29}-\ref{line30}).
A Dense layer with the \textit{ReLU} activation function is added to the Propagation Branch (Line~\ref{line31}) to obtain the Propagation Embedding. 

\begin{algorithm*}[!ht]
\small
\SetAlgoNoLine
\DontPrintSemicolon
\newcommand{\hrulealg}[0]{\vspace{1mm} \hrule \vspace{1mm}}

\SetKwInOut{Input}{Input}
\SetKwInOut{Output}{Output}
\SetKwFor{Loop}{Loop}{}{EndLoop}
\Input{dataset $X=\{ <t_i, v_i, c_i> | i=\overline{1, n} \}$ \newline
    the graph $G=\{ V, E \}$ \newline
    configuration Text Branch $<bidirection, reccurent>$   \newline
    word embedding model for the Text Branch $we\_model$ \newline
    node embedding model for the Propagation Branch $ne\_model$
    }
\Output{GETAE model $m$}

\emph{$T \gets \emptyset$}\label{line01} \atcp{text list}
\emph{$V \gets \emptyset$}\label{line02} \atcp{nodes list}
\emph{$C \gets \emptyset$}\label{line03} \atcp{label list}

\ForEach{$<t, v, c> \in X$}{\label{line04}
    \emph{$T \gets t \cup \{t\}$}\label{line05} \atcp{populate the text list}
    \emph{$V \gets V \cup \{v\}$}\label{line06} \atcp{populate the node list}
    \emph{$C \gets C \cup \{c\}$}\label{line07} \atcp{populate the label list}
}

\emph{$D \gets getDocument2Token(T)$}\label{line08} \atcp{get document to token matrix}
\emph{$W \gets getWordEmbeding(T, we\_model)$}\label{line09} \atcp{train the word embedding}
\emph{$N \gets getNodeEmbeding(G, V, ne\_model)$}\label{line10} \atcp{train the node embedding}

\tcp{==================== Text Branch ========================================}

\emph{$i\_text \gets InputLayer(D, W)$}\label{line11} \atcp{input for the Text Branch}
\emph{$text\_branch.addLayer(i\_text)$}\label{line12}  \atcp{Text Branch implementation}
\emph{$text\_branch.addLayer(WordEmbedding)$}\label{line13} \atcp{Word Embedding layer}

\tcp{==================== Text Branch [Bi]RNN Layer ==========================}
\If{$bidirection = True$}{\label{line14} 
    \If{$reccurent = RNN$}{\label{line15}
        \emph{$text\_branch.addLayer(BiRNN())$}\label{line16}
    }
    \ElseIf{$reccurent = LSTM$}{\label{line17}
        \emph{$text\_branch.addLayer(BiLSTM())$}\label{line18}
    }
    \Else{\label{line19}
         \emph{$text\_branch.addLayer(BiGRU())$}\label{line20}
    }
}
\Else{\label{line21}
    \If{$reccurent = RNN$}{\label{line22}
        \emph{$text\_branch.addLayer(RNN())$}\label{line23}
    }
    \ElseIf{$reccurent = LSTM$}{\label{line24}
        \emph{$text\_branch.addLayer(LSTM())$}\label{line25}
    }
    \Else{\label{line26}
         \emph{$text\_branch.addLayer(GRU())$}\label{line27}
    }
}

\tcp{==================== Propagation Branch =================================}
\emph{$i\_propagation \gets InputLayer(V, N)$}\label{line28} \atcp{input for the Propagation Branch}
\emph{$propagation\_branch.addLayer(i\_propagation)$}\label{line29} \atcp{Propagation Branch implementation}
\emph{$propagation\_branch.addLayer(NodeEmbedding())$}\label{line30} \atcp{Node Embedding layer}
\emph{$propagation\_branch.addLayer(Dense(ReLU))$}\label{line31}  \atcp{Dense layer}

\tcp{==================== Ensemble model and training ========================}

\emph{$ensamble \gets Concatenate(text\_branch, propagation\_branch) $}\label{line32}\atcp{concatenation layer}
\emph{$ensamble.addLayer(Dense(ReLU))$}\label{line33} \atcp{Propagation-Enhanced Content Embedding}
\emph{$o\_layer \gets ensamble.addLayer(Dense(softmax))$}\label{line34} \atcp{model output}
\emph{$m \gets Model(inputs=[i\_text, i\_propagation], output=o\_layer)$}\label{line35} \atcp{model definition}

\emph{$m.train([(D, W), (V, N)], c) $}\label{line36}  \atcp{model training}

\Return{$m$}\;\label{line37}
\caption{GETAE}
\label{alg:getae}
\end{algorithm*}\DecMargin{1em}

The ensemble is created by concatenating the Text and the Propagation Branches (Line~\ref{line32}).
The new Propagation-Enhanced Content Embedding is created by adding a Dense layer with the \textit{ReLU} activation function (Line~\ref{line33}).
The output layer (Line~\ref{line34}) uses as input the Propagation-Enhanced Content Embedding and a Dense layer with the \textit{softmax} activation function to obtain the final prediction.
The model is defined using as input the input layers from the Text and Propagation Branches (Line~\ref{line35}) and as output the last Dense layer (Line~\ref{line34}).
Finally, the model is trained (Line~\ref{line36}) and returned (Line~\ref{line37}).
Subsection~\ref{ssec:ensamble} explains in detail the GETAE's ensemble.

\section{Experimental Results}\label{sec:results}

\subsection{Datasets}

We use \href{https://www.dropbox.com/s/7ewzdrbelpmrnxu/rumdetect2017.zip?dl=0&file_subpath=%2Frumor_detection_acl2017%2Ftwitter15}{Twitter15} and \href{https://www.dropbox.com/s/7ewzdrbelpmrnxu/rumdetect2017.zip?dl=0&file_subpath=%2Frumor_detection_acl2017%2Ftwitter16}{Twitter16} datasets~\cite{Ma2017} for our experimental validation.
Both datasets contain Twitter data separated into: 
1) the \textit{source tweets} with the tweet ID and its text content, 
2) the \textit{label} that assigns a class to each tweet ID, and 
3) the \textit{propagation tree} of each tweet.

\subsubsection{Textual Data Analysis}

\begin{figure}[!htbp]
\centering
\includegraphics[width=1\columnwidth]{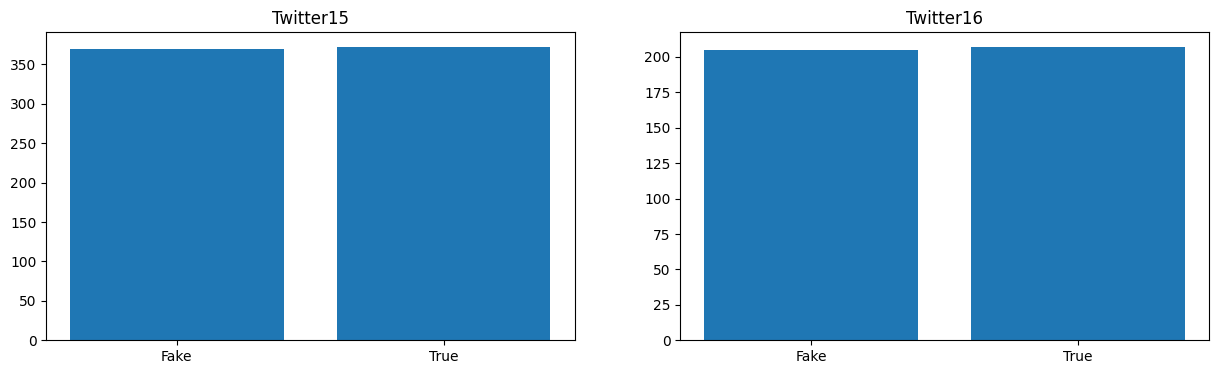}
\caption{Class distribution}
\label{fig:balance}
\end{figure}

Both datasets are labeled using 4 classes: false, true, non-rumor, and unverified.
Because we are building binary classification models, we only used the source tweets labeled as $true$ and $false$ for training. 
Figure~\ref{fig:balance} presents the class distribution for these two labels.
We observe that both datasets are fairly balanced.

\begin{figure}[!htbp]
\centering
\includegraphics[width=1\columnwidth]{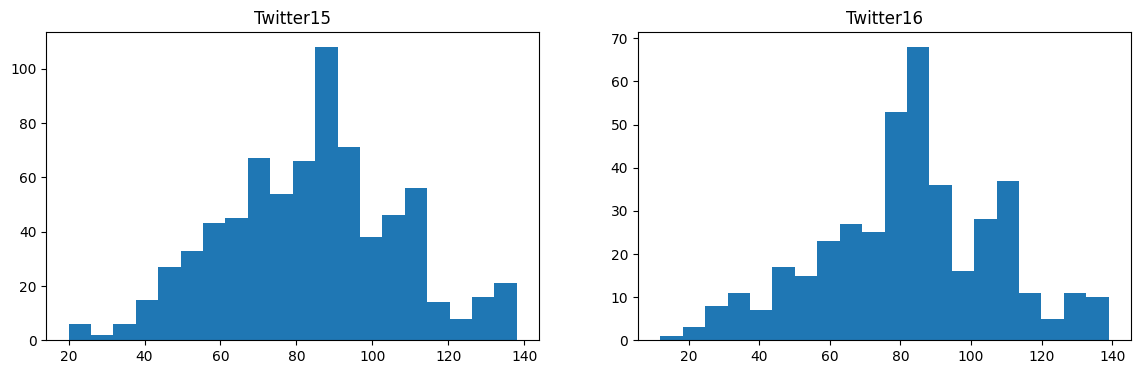}
\caption{Length distribution}
\label{fig:lend}
\end{figure}

\begin{figure}[!htbp]
\centering
\includegraphics[width=1\columnwidth]{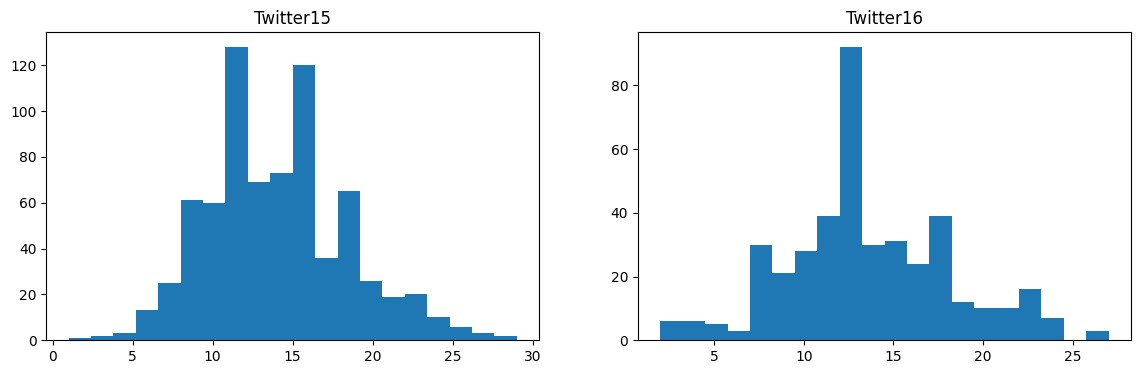}
\caption{Number of words distribution}
\label{fig:nowordsd}
\end{figure}

Figure~\ref{fig:lend} represents the length distribution of tweets in both datasets.
We observe that the most probable length is around 80-100 characters.
Figure~\ref{fig:nowordsd} presents the histogram representing the frequency of the number of words.
Both datasets have most tweets in 10-15 words range and a maximum word count under 30 words.
This highlights the nature of tweets as short texts with several words and a low number of characters.

As both datasets are balanced, we use appropriate evaluation metrics~\cite{Truica2017} to evaluate the trained models, i.e., Accuracy, Precision, Recall, and F1-Score.

\subsubsection{Network Data Analysis}
\begin{figure}[!htbp]
\centering
\includegraphics[width=0.49\columnwidth]{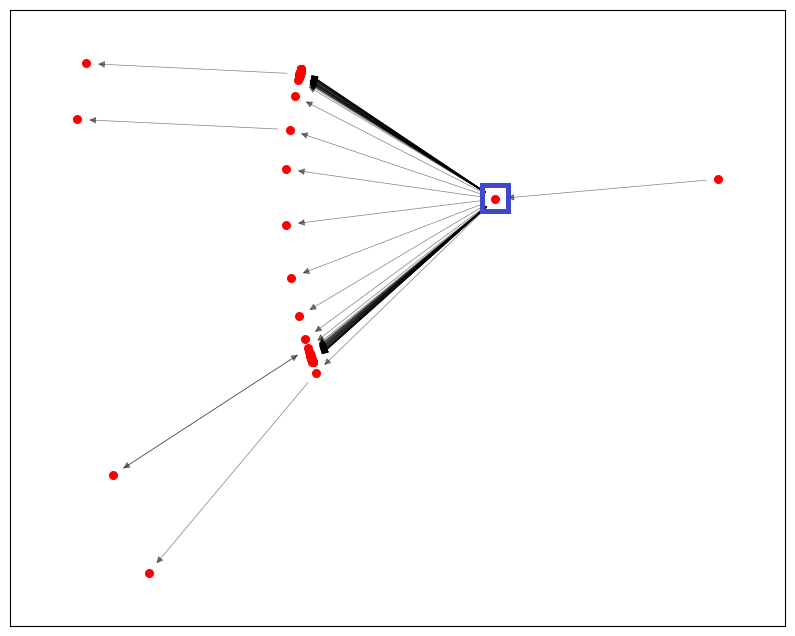}
\includegraphics[width=0.49\columnwidth]{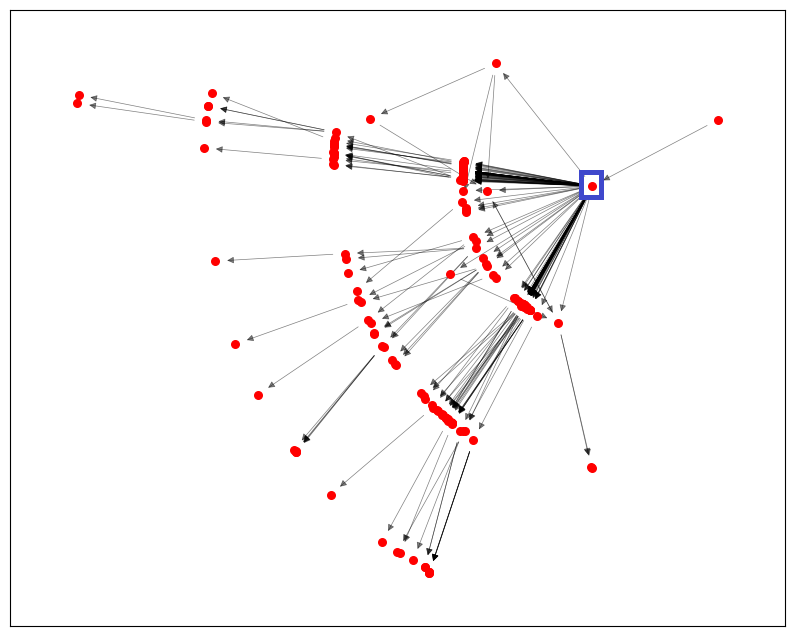}
\caption{Examples of propagation trees}
\label{fig:trees}
\end{figure}

For both Twitter15 and Twitter16 datasets, each propagation tree is encoded using a list of edges from parent nodes to child nodes. 
The relationship between a child and a parent node represents retweets and mentions of the source tweet, only for the neighbor of the root node.
Figure~\ref{fig:trees} presents some examples of propagation trees to better visualize the actual information spread.

The graphs are directed with the orientation specified in the dataset.
Every row in the edge file is a connection from a parent node to a child node. Nodes are represented as tuples with three components: the $user_{id}$ of the Twitter account, the $tweet_{id}$ of the tweet post, and the $time$ in minutes passed since the first node posted (source node).
Equation~\eqref{eq:edge} shows the edge information from the tree file:

\begin{equation}\label{eq:edge}
        (parent_{id}, tweet_{id}, t_{tweet}) \rightarrow (child_{id}, reply_{id}, t_{reply})    
\end{equation}

\begin{figure}[!htbp]
\centering
\subfloat[Twitter15]{\includegraphics[width=0.49\columnwidth]{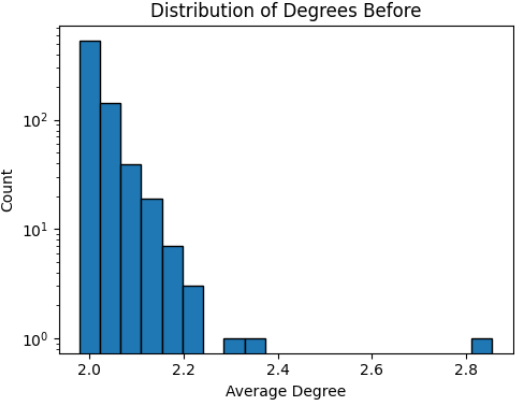}}
\subfloat[Twitter16]{\includegraphics[width=0.49\columnwidth]{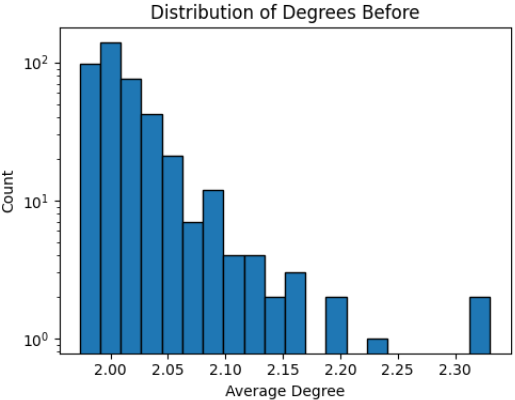}}
\caption{Degree Distribution}
\label{fig:deg_dist}
\end{figure}

Figure~\ref{fig:deg_dist} presents the graphs' Degree Distributions using histograms with a logarithmic scale.
For the Twitter15 dataset, the total number of graphs is 742 and we compute the average degree for each of these graphs. 
We observe that most graphs have a degree distribution around 2, with some of them even over 2.8.
The same behavior is present in the Twitter16 dataset as well, where we observe that most graphs have an average distribution degree over 2.

\begin{figure}[!htbp]
\centering
\subfloat[Twitter15]{\includegraphics[width=1\columnwidth]{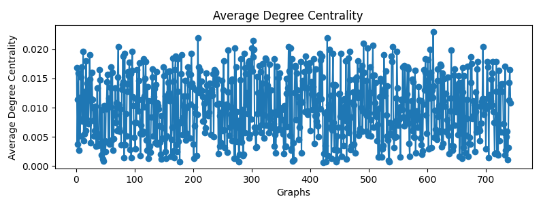}}
\hfill
\subfloat[Twitter16]{\includegraphics[width=1\columnwidth]{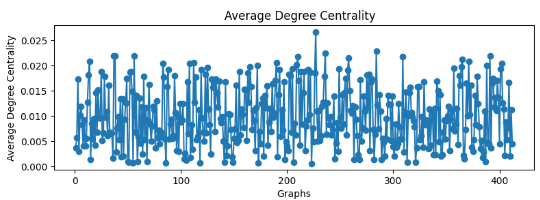}}
\caption{Degree Centrality}
\label{fig:deg_centrality}
\end{figure}

Figure \ref{fig:deg_centrality} presents plots that represent the average Degree Centrality, to measure the immediate risk for infection for a node by counting their neighbors.
We observe that both datasets have similar centrality degrees, thus, the nodes have a similar number of neighbors.

\subsection{Experimental Setup and Implementation}

We conducted experiments on Twitter15 and Twitter16 datasets creating word embeddings, node embeddings, and deep learning models that combine this data.
The text preprocessing for training word embeddings consists of removing
1) URLs, 
2) double spaces, and
3) punctuations.
To vectorize the text, we train a Wor2Vec Skip-Gram model. 
For BERT and BERTweet, we use the pre-trained models from HuggingFace: 'ber-base-uncased' and 'vinai/bertweet-base'.
For Transformer embeddings, we applied the tokenizer with parameters presented in Table~\ref{tb:hyperbert}.
We pass the input\_ids and attention\_mask tensors to the BERT models to obtain the transformer embeddings. 
For BERTweet, the only difference was loading a different tokenizer and model from huggingface.
Table \ref{tb:hyperwe} specifies the hyperparameters used for obtaining node embeddings with \textsc{Node2Vec} and \textsc{DeepWalk}. 
Table~\ref{tb:hyperlayers} presents GETAE's deep learning ensemble architecture hyperparameters. 

\begin{table}[!htbp]
\caption{Embeddings hyperparameters}
\label{tb:hyperwe}
\resizebox{\columnwidth}{!}{
\begin{tabular}{lrrrrr}
\hline
\textbf{Embeddings} & \textbf{Window} & \textbf{Min Count} & \textbf{Walk Length} & \textbf{Learning Rate} & \textbf{Epochs} \\ \hline
\textsc{Word2Vec} & 10 & 4 & - & 0.025 & 5 \\ \hline
\textsc{Node2Vec} & 10 & 1 & 10 & 0.05 & 1 \\ \hline
\textsc{DeepWalk} & 5 & 1 & 10 & 0.05 & 1 \\ \hline
\end{tabular}
}
\end{table}

\begin{table}[!htbp]
\caption{BERT hyperparameters}
\label{tb:hyperbert}
\resizebox{\columnwidth}{!}{
\begin{tabular}{lrrrrr}
\hline
\textbf{Tokenizer} & \textbf{add\_special\_tokens} & \textbf{padding} & \textbf{truncation} & \textbf{max\_length} \\ \hline
BERT     & TRUE & TRUE & TRUE & 128 \\ \hline
\textsc{BERTweet} & TRUE & TRUE & TRUE & 128 \\ \hline
\end{tabular}
}
\end{table}

\begin{table}[!htbp]
\centering
\caption{GETAE's layers hyperparameters}
\label{tb:hyperlayers}
\resizebox{\columnwidth}{!}{
\begin{tabular}{lrrl}
\hline
\textbf{Layer} & \textbf{Units} & \textbf{Dropout} & \textbf{Activation Function} \\ \hline
RNN  & 64 & 0.2 & Tanh (Hyperbolic Tangent) \\ \hline
\textsc{BiRNN}  & 128 & 0.2 & Tanh (Hyperbolic Tangent) \\ \hline
GRU  & 64 & 0.2 &  Tanh (Hyperbolic Tangent) \\ \hline
\textsc{BiGRU}  & 128 & 0.2 & Tanh (Hyperbolic Tangent) \\ \hline
LSTM & 64 & 0.2 &  Tanh (Hyperbolic Tangent) \\ \hline
\textsc{BiLSTM} & 128 & 0.2 &  Tanh (Hyperbolic Tangent) \\ \hline
Text Dense & 32 & - & ReLU (Rectified Linear Unit) \\ \hline
Graph Dense & 32 & - & ReLU (Rectified Linear Unit) \\ \hline
Output Dense & 1 & - & Sigmoid \\ \hline
\end{tabular}
}

\end{table}

We implemented the preprocessing elements of GETAE in Python version 3.10.
For text preprocessing, we use the \href{https://www.nltk.org/}{nltk}~\cite{Bird2009} package.
For training the \textsc{Word2Vec} word embeddings, we use the \href{https://radimrehurek.com/gensim/}{gensim}~\cite{Rehurek2010} package. 
For training the node embeddings, we use the \href{https://github.com/eliorc/node2vec}{node2vec} and \href{https://github.com/benedekrozemberczki/karateclub}{karateclub}~\cite{Rozemberczki2020} packages for training the embeddings with \textsc{Node2Vec} and \textsc{DeepWalk}, respectively.
We implemented the GETAE architecture and trained the models using \href{https://www.tensorflow.org/}{TensorFlow} and \href{https://keras.io/}{Keras}.
The code is freely available on GitHub at \url{https://github.com/DS4AI-UPB/GETAE}.

\subsection{Classification and Ablation Results}\label{ssec:class}

For this set of experiments, we use an 80-20 train-test ratio.
We trained the multiple models using the GETAE architecture for 30 epochs, except for the models that employ the \textsc{Word2Vec} word embeddings which were trained for only 8 epochs to avoid overfitting.
For training the different GETAE models, we used a learning rate of $0.001$. 
The training process for the models was carried out on Google Colab, with no hardware acceleration.

All experiments use k-fold cross-validation with $k=10$ with the ratio 80-20 for the train and test sets.
The tables present the mean and standard deviation for each metric, using the individual results obtained at each fold during the k-fold cross-validation.

Tables~\ref{tb:we15} and~\ref{tb:we16} present the results obtained on the test set after training the GETAE models on Twitter15 and Twitter16, respectively.
We use three types of word embeddings (i.e., \textsc{Word2Vec} Skip-Gram, BERT, and \textsc{BERTweet}) and two types of node embeddings (i.e., \textsc{Node2Vec} and \textsc{DeepWalk}).
These tables also present two types of ablation testing for the GETAE architecture: 
1) adding or removing the Propagation Branch,
2) changing the RNN Layer for the Text Branch.
The Text Branch uses six different recurrent neural network layers for ablation testing: RNN, \textsc{BiRNN}, GRU, \textsc{BiGRU}, LSTM, and \textsc{BiLSTM}. 

The evaluation of the models obtained by training the GETAE architecture includes performance metrics such as Accuracy, Precision, Recall, and F1-Score, measured on both Twitter15 and Twitter16 datasets. 
The experiments are conducted with and without the Network Branch for ablation testing.
The hyperparameters for \textsc{Node2Vec} are: $p = 1$, $q = 1$, $d = 100$, where $p$ is the return hyperparameter, $q$ is in-out hyperparameter, and $d$ is the embeddings' dimensions.
For \textsc{DeepWalk} node embeddings, the only hyperparameter tuned was the dimensionality of the embedding space, set to $d = 100$.

For the Twitter15 dataset, the best configuration for GETAE by metrics is a tie between configurations (BERT, \textsc{Node2Vec}, \textsc{BiLSTM}) and (\textsc{BERTweet}, \textsc{DeepWalk}, \textsc{BiRNN}). 
When comparing these two models with other state-of-the-art ones, we choose the (BERT, \textsc{Node2Vec}, \textsc{BiLSTM}) configuration due to a higher F1-Score.
\textsc{BERTweet} should also be mentioned as having promising results, especially when employing the \textsc{DeepWalk} node embedding and \textsc{BiRNN} as the recurrent layer.

\begin{table*}[!htbp]
\caption{GETAE ablation Testing on Twitter15 (\textsc{Node2Vec} $p = 1$, $q = 1$, $dimensions = 100$ for  both \textsc{Node2Vec} and \textsc{DeepWalk}.)}
\label{tb:we15}
\centering
\resizebox{2\columnwidth}{!}{
\begin{tabular}{|l|l|c|c|r|r|r|r|r|r|r|r|}
\hline
\multirow{2}{*}{\textbf{\begin{tabular}[c]{@{}c@{}}Word\\ Embedding\end{tabular}}} & \multirow{2}{*}{\textbf{\begin{tabular}[c]{@{}c@{}}Network\\ Embedding\end{tabular}}}
& \multirow{2}{*}{\textbf{\begin{tabular}[c]{@{}c@{}}Text\\ Branch\end{tabular}}}& \multirow{2}{*}{\textbf{\begin{tabular}[c]{@{}c@{}}Propagation\\ Branch\end{tabular}}}  & 
\multicolumn{4}{c|}{\textbf{RNN Layer}}   & \multicolumn{4}{c|}{\textbf{\textsc{BiRNN} Layer}}                      \\ \cline{5-12}

& & & & \textbf{Accuracy} & \textbf{Precision} & \textbf{Recall} & \textbf{F1-Score} & \textbf{Accuracy} & \textbf{Precision} & \textbf{Recall} & \textbf{F1-Score} \\ \hline

\multirow{3}{*}{\textsc{\textbf{Word2Vec}}}    & \textbf{N/A}                 & \checkmark & N/A        & 0.749 $\pm$ 0.039 & 0.764 $\pm$ 0.036 & 0.750 $\pm$ 0.042 & 0.745 $\pm$ 0.043 & 0.754 $\pm$ 0.027 & 0.763 $\pm$ 0.027 & 0.754 $\pm$ 0.028 & 0.751 $\pm$ 0.027 \\ \cline{2-12}
                                               & \textsc{\textbf{Node2Vec}}   & \checkmark & \checkmark & 0.651 $\pm$ 0.028 & 0.683 $\pm$ 0.025 & 0.660 $\pm$ 0.027 & 0.642 $\pm$ 0.041 & 0.664 $\pm$ 0.022 & 0.698 $\pm$ 0.042 & 0.675 $\pm$ 0.025 & 0.656 $\pm$ 0.023 \\ \cline{2-12}
                                               & \textsc{\textbf{DeepWalk}}   & \checkmark & \checkmark & 0.705 $\pm$ 0.025 & 0.720 $\pm$ 0.023 & 0.711 $\pm$ 0.023 & 0.703 $\pm$ 0.025 & 0.703 $\pm$ 0.043 & 0.729 $\pm$ 0.028 & 0.711 $\pm$ 0.039 & 0.698 $\pm$ 0.049 \\ \hline

\multirow{3}{*}{\textsc{\textbf{BERT}}}        & \textbf{N/A}                 & \checkmark & N/A        & 0.790 $\pm$ 0.036 & 0.791 $\pm$.0.034 & 0.788 $\pm$ 0.035 & 0.784 $\pm$ 0.037 & 0.797 $\pm$ 0.033 & 0.800 $\pm$.0.034 & 0.797 $\pm$ 0.033 & 0.794 $\pm$ 0.034 \\ \cline{2-12}
                                               & \textsc{\textbf{Node2Vec}}   & \checkmark & \checkmark & 0.766 $\pm$ 0.070 & 0.800 $\pm$ 0.035 & 0.769 $\pm$ 0.066 & 0.755 $\pm$ 0.090 & 0.800 $\pm$ 0.028 & 0.810 $\pm$ 0.024 & 0.801 $\pm$ 0.029 & 0.797 $\pm$ 0.029 \\ \cline{2-12}
                                               & \textsc{\textbf{DeepWalk}}   & \checkmark & \checkmark & 0.736 $\pm$ 0.063 & 0.773 $\pm$ 0.044 & 0.738 $\pm$ 0.060 & 0.725 $\pm$ 0.071 & 0.795 $\pm$ 0.031 & 0.802 $\pm$ 0.032 & 0.796 $\pm$ 0.030 & 0.791 $\pm$ 0.031 \\ \hline

\multirow{3}{*}{\textsc{\textbf{BERTweet}}}    & \textbf{N/A}                 & \checkmark & N/A        & 0.735 $\pm$ 0.084 & 0.780 $\pm$ 0.024 & 0.736 $\pm$ 0.086 & 0.714 $\pm$ 0.128 & 0.792 $\pm$ 0.059 & 0.824 $\pm$ 0.026 & 0.797 $\pm$ 0.053 & 0.787 $\pm$ 0.064 \\ \cline{2-12}
                                               & \textsc{\textbf{Node2Vec}}   & \checkmark & \checkmark & 0.701 $\pm$ 0.097 & 0.773 $\pm$ 0.042 & 0.698 $\pm$ 0.095 & 0.664 $\pm$ 0.139 & 0.790 $\pm$ 0.050 & 0.812 $\pm$ 0.032 & 0.790 $\pm$ 0.044 & 0.783 $\pm$ 0.054 \\ \cline{2-12}
                                               & \textsc{\textbf{DeepWalk}}   & \checkmark & \checkmark & 0.736 $\pm$ 0.075 & 0.772 $\pm$ 0.039 & 0.725 $\pm$ 0.084 & 0.711 $\pm$ 0.113 & 0.827 $\pm$ 0.041 & \textbf{0.839 $\pm$ 0.031} & \textbf{0.827 $\pm$ 0.044} & 0.823$\pm$0.046 \\ \hline

\multirow{2}{*}{\textbf{\begin{tabular}[c]{@{}c@{}}Word\\ Embedding\end{tabular}}} & \multirow{2}{*}{\textbf{\begin{tabular}[c]{@{}c@{}}Network\\ Embedding\end{tabular}}}
& \multirow{2}{*}{\textbf{\begin{tabular}[c]{@{}c@{}}Text\\ Branch\end{tabular}}}& \multirow{2}{*}{\textbf{\begin{tabular}[c]{@{}c@{}}Propagation\\ Branch\end{tabular}}}  & 
\multicolumn{4}{c|}{\textbf{GRU Layer}}   & \multicolumn{4}{c|}{\textbf{\textsc{BiGRU} Layer}}                      \\ \cline{5-12}
& & & & \textbf{Accuracy} & \textbf{Precision} & \textbf{Recall} & \textbf{F1-Score} & \textbf{Accuracy} & \textbf{Precision} & \textbf{Recall} & \textbf{F1-Score} \\ \hline

\multirow{3}{*}{\textsc{\textbf{Word2Vec}}}    & \textbf{N/A}                 & \checkmark & N/A        & 0.481 $\pm$ 0.020 & 0.240 $\pm$ 0.010 & 0.500 $\pm$ 0.000 & 0.324 $\pm$ 0.008 & 0.631 $\pm$ 0.043 & 0.696 $\pm$ 0.047 & 0.636 $\pm$ 0.044 & 0.603 $\pm$ 0.066 \\ \cline{2-12}
                                               & \textsc{\textbf{Node2Vec}}   & \checkmark & \checkmark & 0.485 $\pm$ 0.027 & 0.351 $\pm$ 0.150 & 0.510 $\pm$ 0.017 & 0.371 $\pm$ 0.085 & 0.614 $\pm$ 0.029 & 0.702 $\pm$ 0.053 & 0.627 $\pm$ 0.024 & 0.584 $\pm$ 0.052 \\ \cline{2-12}
                                               & \textsc{\textbf{DeepWalk}}   & \checkmark & \checkmark & 0.485 $\pm$ 0.033 & 0.491 $\pm$ 0.034 & 0.492 $\pm$ 0.033 & 0.469 $\pm$ 0.044 & 0.625 $\pm$ 0.027 & 0.656 $\pm$ 0.045 & 0.631 $\pm$ 0.034 & 0.614 $\pm$ 0.031 \\ \hline
                                               
\multirow{3}{*}{\textsc{\textbf{BERT}}}        & \textbf{N/A}                 & \checkmark & N/A        & 0.764 $\pm$ 0.031 & 0.772 $\pm$.0.027 & 0.771 $\pm$ 0.025 & 0.761 $\pm$ 0.029 & 0.780 $\pm$ 0.018 & 0.784 $\pm$.0.018 & 0.783 $\pm$ 0.017 & 0.779 $\pm$ 0.018 \\ \cline{2-12}
                                               & \textsc{\textbf{Node2Vec}}   & \checkmark & \checkmark & 0.757 $\pm$ 0.044 & 0.765 $\pm$ 0.041 & 0.760 $\pm$ 0.046 & 0.755 $\pm$ 0.045 & 0.768 $\pm$ 0.030 & 0.769 $\pm$ 0.025 & 0.766 $\pm$ 0.027 & 0.764 $\pm$ 0.029 \\ \cline{2-12}
                                               & \textsc{\textbf{DeepWalk}}   & \checkmark & \checkmark & 0.751 $\pm$ 0.036 & 0.751 $\pm$ 0.036 & 0.750 $\pm$ 0.036 & 0.749 $\pm$ 0.035 & 0.776 $\pm$ 0.015 & 0.776 $\pm$ 0.017 & 0.775 $\pm$ 0.014 & 0.773 $\pm$ 0.016 \\ \hline
                                               
\multirow{3}{*}{\textsc{\textbf{BERTweet}}}    & \textbf{N/A}                 & \checkmark & N/A        & 0.686 $\pm$ 0.044 & 0.734 $\pm$ 0.038 & 0.688 $\pm$ 0.040 & 0.667 $\pm$ 0.060 & 0.742 $\pm$ 0.038 & 0.753 $\pm$ 0.030 & 0.746 $\pm$ 0.036 & 0.742 $\pm$ 0.038 \\ \cline{2-12}
                                               & \textsc{\textbf{Node2Vec}}   & \checkmark & \checkmark & 0.694 $\pm$ 0.072 & 0.757 $\pm$ 0.038 & 0.695 $\pm$ 0.058 & 0.670 $\pm$ 0.091 & 0.731 $\pm$ 0.023 & 0.752 $\pm$ 0.033 & 0.735 $\pm$ 0.019 & 0.727 $\pm$ 0.021 \\ \cline{2-12}
                                               & \textsc{\textbf{DeepWalk}}   & \checkmark & \checkmark & 0.712 $\pm$ 0.044 & 0.745 $\pm$ 0.034 & 0.713 $\pm$ 0.049 & 0.701 $\pm$ 0.057 & 0.748 $\pm$ 0.029 & 0.754 $\pm$ 0.029 & 0.750 $\pm$ 0.028 & 0.747 $\pm$ 0.028 \\ \hline

\multirow{2}{*}{\textbf{\begin{tabular}[c]{@{}c@{}}Word\\ Embedding\end{tabular}}} & \multirow{2}{*}{\textbf{\begin{tabular}[c]{@{}c@{}}Network\\ Embedding\end{tabular}}}
& \multirow{2}{*}{\textbf{\begin{tabular}[c]{@{}c@{}}Text\\ Branch\end{tabular}}}& \multirow{2}{*}{\textbf{\begin{tabular}[c]{@{}c@{}}Propagation\\ Branch\end{tabular}}}  & 
\multicolumn{4}{c|}{\textbf{LSTM}}   & \multicolumn{4}{c|}{\textbf{\textsc{BiLSTM} Layer}}                      \\ \cline{5-12}
& & & & \textbf{Accuracy} & \textbf{Precision} & \textbf{Recall} & \textbf{F1-Score} & \textbf{Accuracy} & \textbf{Precision} & \textbf{Recall} & \textbf{F1-Score} \\ \hline

\multirow{3}{*}{\textsc{\textbf{Word2Vec}}}    & \textbf{N/A}                 & \checkmark & N/A        & 0.667 $\pm$ 0.051 & 0.732 $\pm$ 0.046 & 0.668 $\pm$ 0.045 & 0.643 $\pm$ 0.066 & 0.669 $\pm$ 0.038 & 0.712 $\pm$ 0.042 & 0.673 $\pm$ 0.037 & 0.653 $\pm$ 0.045 \\ \cline{2-12}
                                               & \textsc{\textbf{Node2Vec}}   & \checkmark & \checkmark & 0.623 $\pm$ 0.040 & 0.649 $\pm$ 0.053 & 0.630 $\pm$ 0.040 & 0.617 $\pm$ 0.042 & 0.590 $\pm$ 0.037 & 0.674 $\pm$ 0.063 & 0.605 $\pm$ 0.039 & 0.556 $\pm$ 0.056 \\ \cline{2-12}
                                               & \textsc{\textbf{DeepWalk}}   & \checkmark & \checkmark & 0.642 $\pm$ 0.028 & 0.679 $\pm$ 0.034 & 0.653 $\pm$ 0.028 & 0.632 $\pm$ 0.030 & 0.647 $\pm$ 0.023 & 0.688 $\pm$ 0.035 & 0.657 $\pm$ 0.021 & 0.636 $\pm$ 0.036 \\ \hline 
                                               
\multirow{3}{*}{\textsc{\textbf{BERT}}}        & \textbf{N/A}                 & \checkmark & N/A        & 0.756 $\pm$ 0.049 & 0.788 $\pm$.0.019 & 0.758 $\pm$ 0.043 & 0.746 $\pm$ 0.054 & 0.799 $\pm$ 0.030 & 0.805 $\pm$.0.025 & 0.796 $\pm$ 0.034 & 0.794 $\pm$ 0.036 \\ \cline{2-12}
                                               & \textsc{\textbf{Node2Vec}}   & \checkmark & \checkmark & 0.783 $\pm$ 0.045 & 0.805 $\pm$ 0.031 & 0.775 $\pm$ 0.049 & 0.773 $\pm$ 0.053 & \textbf{0.827 $\pm$ 0.042} & 0.831 $\pm$ 0.035 & 0.827 $\pm$ 0.040 & \textbf{0.825 $\pm$ 0.042} \\ \cline{2-12}
                                               & \textsc{\textbf{DeepWalk}}   & \checkmark & \checkmark & 0.764 $\pm$ 0.050 & 0.782 $\pm$ 0.035 & 0.760 $\pm$ 0.050 & 0.757 $\pm$ 0.059 & 0.814 $\pm$ 0.043 & 0.818 $\pm$ 0.041 & 0.815 $\pm$ 0.042 & 0.812 $\pm$ 0.044 \\ \hline
                                               
\multirow{3}{*}{\textsc{\textbf{BERTweet}}}        & \textbf{N/A}                 & \checkmark & N/A        & 0.651 $\pm$ 0.105 & 0.758 $\pm$ 0.040 & 0.666 $\pm$ 0.096 & 0.611 $\pm$ 0.139 & 0.748 $\pm$ 0.049 & 0.777 $\pm$ 0.029 & 0.752 $\pm$ 0.042 & 0.741 $\pm$ 0.053 \\ \cline{2-12}
                                               & \textsc{\textbf{Node2Vec}}   & \checkmark & \checkmark & 0.708 $\pm$ 0.078 & 0.783 $\pm$ 0.028 & 0.702 $\pm$ 0.078 & 0.675 $\pm$ 0.117 & 0.758 $\pm$ 0.047 & 0.787 $\pm$ 0.035 & 0.759 $\pm$ 0.049 & 0.749 $\pm$ 0.058 \\ \cline{2-12}
                                               & \textsc{\textbf{DeepWalk}}   & \checkmark & \checkmark & 0.732 $\pm$ 0.074 & 0.762 $\pm$ 0.037 & 0.735 $\pm$ 0.062 & 0.720 $\pm$ 0.093 & 0.770 $\pm$ 0.083 & 0.815 $\pm$ 0.029 & 0.774 $\pm$ 0.077 & 0.757 $\pm$ 0.103 \\ \hline
                                               
\end{tabular}
}
\end{table*}

\begin{table*}[!ht]
\centering
\caption{GETAE ablation Testing on Twitter16 (\textsc{Node2Vec} $p = 1$, $q = 1$, $dimensions = 100$ for  both \textsc{Node2Vec} and \textsc{DeepWalk}.)}
\label{tb:we16}

\resizebox{2\columnwidth}{!}{
\begin{tabular}{|l|l|c|c|r|r|r|r|r|r|r|r|}
\hline
\multirow{2}{*}{\textbf{\begin{tabular}[c]{@{}c@{}}Word\\ Embedding\end{tabular}}} & \multirow{2}{*}{\textbf{\begin{tabular}[c]{@{}c@{}}Network\\ Embedding\end{tabular}}}
& \multirow{2}{*}{\textbf{\begin{tabular}[c]{@{}c@{}}Text\\ Branch\end{tabular}}}& \multirow{2}{*}{\textbf{\begin{tabular}[c]{@{}c@{}}Propagation\\ Branch\end{tabular}}}  & 
\multicolumn{4}{c|}{\textbf{RNN Layer}}   & \multicolumn{4}{c|}{\textbf{\textsc{BiRNN} Layer}}                      \\ \cline{5-12}

& & & & \textbf{Accuracy} & \textbf{Precision} & \textbf{Recall} & \textbf{F1-Score} & \textbf{Accuracy} & \textbf{Precision} & \textbf{Recall} & \textbf{F1-Score} \\ \hline

\multirow{3}{*}{\textsc{\textbf{Word2Vec}}}    & \textbf{N/A}                 & \checkmark & N/A        & 0.705 $\pm$ 0.081 & 0.716 $\pm$ 0.075 & 0.705 $\pm$ 0.072 & 0.697 $\pm$ 0.080 & 0.699 $\pm$ 0.066 & 0.712 $\pm$ 0.071 & 0.698 $\pm$ 0.063 & 0.691 $\pm$ 0.068 \\ \cline{2-12}
                                               & \textsc{\textbf{Node2Vec}}   & \checkmark & \checkmark & 0.647 $\pm$ 0.048 & 0.685 $\pm$ 0.074 & 0.646 $\pm$ 0.048 & 0.627 $\pm$ 0.053 & 0.728 $\pm$ 0.085 & 0.756 $\pm$ 0.081 & 0.735 $\pm$ 0.087 & 0.719 $\pm$ 0.095 \\ \cline{2-12}
                                               & \textsc{\textbf{DeepWalk}}   & \checkmark & \checkmark & 0.702 $\pm$ 0.060 & 0.722 $\pm$ 0.067 & 0.707 $\pm$ 0.061 & 0.699 $\pm$ 0.062 & 0.695 $\pm$ 0.062 & 0.726 $\pm$ 0.057 & 0.698 $\pm$ 0.059 & 0.687 $\pm$ 0.070 \\ \hline

\multirow{3}{*}{\textsc{\textbf{BERT}}}        & \textbf{N/A}                 & \checkmark & N/A        & 0.848 $\pm$ 0.036 & 0.850 $\pm$ 0.032 & 0.849 $\pm$ 0.036 & 0.847 $\pm$ 0.037 & 0.856 $\pm$ 0.043 & 0.869 $\pm$ 0.029 & 0.863 $\pm$ 0.035 & 0.854 $\pm$ 0.045 \\ \cline{2-12}
                                               & \textsc{\textbf{Node2Vec}}   & \checkmark & \checkmark & \textbf{0.896 $\pm$ 0.025} & \textbf{0.901 $\pm$ 0.026} & \textbf{0.897 $\pm$ 0.028} & \textbf{0.895 $\pm$ 0.025} & 0.880 $\pm$ 0.031 & 0.881 $\pm$ 0.031 & 0.880 $\pm$ 0.031 & 0.879 $\pm$ 0.031 \\ \cline{2-12}
                                               & \textsc{\textbf{DeepWalk}}   & \checkmark & \checkmark & 0.801 $\pm$ 0.072 & 0.836 $\pm$ 0.037 & 0.800 $\pm$ 0.079 & 0.789 $\pm$ 0.089 & 0.851 $\pm$ 0.040 & 0.856 $\pm$ 0.036 & 0.856 $\pm$ 0.032 & 0.850 $\pm$ 0.038 \\ \hline

\multirow{3}{*}{\textsc{\textbf{BERTweet}}}    & \textbf{N/A}                 & \checkmark & N/A        & 0.714 $\pm$ 0.098 & 0.743 $\pm$ 0.162 & 0.695 $\pm$ 0.112 & 0.659 $\pm$ 0.161 & 0.789 $\pm$ 0.097 & 0.835 $\pm$ 0.055 & 0.799 $\pm$ 0.090 & 0.781 $\pm$ 0.107 \\ \cline{2-12}
                                               & \textsc{\textbf{Node2Vec}}   & \checkmark & \checkmark & 0.670 $\pm$ 0.101 & 0.666 $\pm$ 0.208 & 0.655 $\pm$ 0.113 & 0.601 $\pm$ 0.170 & 0.800 $\pm$ 0.099 & 0.847 $\pm$ 0.046 & 0.801 $\pm$ 0.096 & 0.784 $\pm$ 0.125 \\ \cline{2-12}
                                               & \textsc{\textbf{DeepWalk}}   & \checkmark & \checkmark & 0.612 $\pm$ 0.122 & 0.670 $\pm$ 0.233 & 0.621 $\pm$ 0.098 & 0.538 $\pm$ 0.162 & 0.829 $\pm$ 0.033 & 0.842 $\pm$ 0.030 & 0.831 $\pm$ 0.031 & 0.826 $\pm$ 0.034 \\ \hline

\multirow{2}{*}{\textbf{\begin{tabular}[c]{@{}c@{}}Word\\ Embedding\end{tabular}}} & \multirow{2}{*}{\textbf{\begin{tabular}[c]{@{}c@{}}Network\\ Embedding\end{tabular}}}
& \multirow{2}{*}{\textbf{\begin{tabular}[c]{@{}c@{}}Text\\ Branch\end{tabular}}}& \multirow{2}{*}{\textbf{\begin{tabular}[c]{@{}c@{}}Propagation\\ Branch\end{tabular}}}  & 
\multicolumn{4}{c|}{\textbf{GRU Layer}}   & \multicolumn{4}{c|}{\textbf{\textsc{BiGRU} Layer}}                      \\ \cline{5-12}
& & & & \textbf{Accuracy} & \textbf{Precision} & \textbf{Recall} & \textbf{F1-Score} & \textbf{Accuracy} & \textbf{Precision} & \textbf{Recall} & \textbf{F1-Score} \\ \hline

\multirow{3}{*}{\textsc{\textbf{Word2Vec}}}    & \textbf{N/A}                 & \checkmark & N/A        & 0.508 $\pm$ 0.062 & 0.303 $\pm$ 0.158 & 0.501 $\pm$ 0.003 & 0.339 $\pm$ 0.029 & 0.590 $\pm$ 0.134 & 0.518 $\pm$ 0.271 & 0.589 $\pm$ 0.121 & 0.478 $\pm$ 0.190 \\ \cline{2-12}
                                               & \textsc{\textbf{Node2Vec}}   & \checkmark & \checkmark & 0.488 $\pm$ 0.070 & 0.438 $\pm$ 0.190 & 0.517 $\pm$ 0.028 & 0.382 $\pm$ 0.095 & 0.502 $\pm$ 0.114 & 0.418 $\pm$ 0.210 & 0.546 $\pm$ 0.085 & 0.407 $\pm$ 0.158 \\ \cline{2-12}
                                               & \textsc{\textbf{DeepWalk}}   & \checkmark & \checkmark & 0.463 $\pm$ 0.027 & 0.409 $\pm$ 0.095 & 0.475 $\pm$ 0.034 & 0.380 $\pm$ 0.040 & 0.509 $\pm$ 0.062 & 0.482 $\pm$ 0.119 & 0.512 $\pm$ 0.056 & 0.451 $\pm$ 0.100 \\ \hline
                                               
\multirow{3}{*}{\textsc{\textbf{BERT}}}        & \textbf{N/A}                 & \checkmark & N/A        & 0.858 $\pm$ 0.038 & 0.868 $\pm$ 0.035 & 0.858 $\pm$ 0.040 & 0.855 $\pm$ 0.039 & 0.837 $\pm$ 0.051 & 0.844 $\pm$ 0.045 & 0.838 $\pm$ 0.048 & 0.835 $\pm$ 0.052 \\ \cline{2-12}
                                               & \textsc{\textbf{Node2Vec}}   & \checkmark & \checkmark & 0.869 $\pm$ 0.049 & 0.876 $\pm$ 0.047 & 0.872 $\pm$ 0.051 & 0.866 $\pm$ 0.050 & 0.858 $\pm$ 0.051 & 0.860 $\pm$ 0.046 & 0.855 $\pm$ 0.053 & 0.855 $\pm$ 0.053 \\ \cline{2-12}
                                               & \textsc{\textbf{DeepWalk}}   & \checkmark & \checkmark & 0.845 $\pm$ 0.031 & 0.849 $\pm$ 0.025 & 0.847 $\pm$ 0.030 & 0.843 $\pm$ 0.032 & 0.857 $\pm$ 0.027 & 0.866 $\pm$ 0.024 & 0.862 $\pm$ 0.031 & 0.856 $\pm$ 0.027 \\ \hline
                                               
\multirow{3}{*}{\textsc{\textbf{BERTweet}}}    & \textbf{N/A}                 & \checkmark & N/A        & 0.693 $\pm$ 0.121 & 0.720 $\pm$ 0.168 & 0.697 $\pm$ 0.120 & 0.653 $\pm$ 0.170 & 0.781 $\pm$ 0.043 & 0.827 $\pm$ 0.024 & 0.788 $\pm$ 0.041 & 0.773 $\pm$ 0.049 \\ \cline{2-12}
                                               & \textsc{\textbf{Node2Vec}}   & \checkmark & \checkmark & 0.672 $\pm$ 0.107 & 0.755 $\pm$ 0.045 & 0.690 $\pm$ 0.094 & 0.644 $\pm$ 0.141 & 0.739 $\pm$ 0.063 & 0.784 $\pm$ 0.026 & 0.751 $\pm$ 0.046 & 0.728 $\pm$ 0.066 \\ \cline{2-12}
                                               & \textsc{\textbf{DeepWalk}}   & \checkmark & \checkmark & 0.684 $\pm$ 0.105 & 0.788 $\pm$ 0.045 & 0.683 $\pm$ 0.094 & 0.638 $\pm$ 0.132 & 0.767 $\pm$ 0.109 & 0.797 $\pm$ 0.067 & 0.774 $\pm$ 0.089 & 0.757 $\pm$ 0.125 \\ \hline

\multirow{2}{*}{\textbf{\begin{tabular}[c]{@{}c@{}}Word\\ Embedding\end{tabular}}} & \multirow{2}{*}{\textbf{\begin{tabular}[c]{@{}c@{}}Network\\ Embedding\end{tabular}}}
& \multirow{2}{*}{\textbf{\begin{tabular}[c]{@{}c@{}}Text\\ Branch\end{tabular}}}& \multirow{2}{*}{\textbf{\begin{tabular}[c]{@{}c@{}}Propagation\\ Branch\end{tabular}}}  & 
\multicolumn{4}{c|}{\textbf{LSTM}}   & \multicolumn{4}{c|}{\textbf{\textsc{BiLSTM} Layer}}                      \\ \cline{5-12}
& & & & \textbf{Accuracy} & \textbf{Precision} & \textbf{Recall} & \textbf{F1-Score} & \textbf{Accuracy} & \textbf{Precision} & \textbf{Recall} & \textbf{F1-Score} \\ \hline

\multirow{3}{*}{\textsc{\textbf{Word2Vec}}}    & \textbf{N/A}                 & \checkmark & N/A        & 0.471 $\pm$ 0.054 & 0.235 $\pm$ 0.025 & 0.500 $\pm$ 0.000 & 0.320 $\pm$ 0.024 & 0.522 $\pm$ 0.116 & 0.403 $\pm$ 0.262 & 0.547 $\pm$ 0.087 & 0.401 $\pm$ 0.150 \\ \cline{2-12}
                                               & \textsc{\textbf{Node2Vec}}   & \checkmark & \checkmark & 0.481 $\pm$ 0.056 & 0.374 $\pm$ 0.134 & 0.506 $\pm$ 0.022 & 0.366 $\pm$ 0.068 & 0.496 $\pm$ 0.046 & 0.491 $\pm$ 0.183 & 0.535 $\pm$ 0.052 & 0.400 $\pm$ 0.083 \\ \cline{2-12}
                                               & \textsc{\textbf{DeepWalk}}   & \checkmark & \checkmark & 0.467 $\pm$ 0.021 & 0.416 $\pm$ 0.098 & 0.478 $\pm$ 0.021 & 0.407 $\pm$ 0.057 & 0.488 $\pm$ 0.050 & 0.440 $\pm$ 0.117 & 0.499 $\pm$ 0.044 & 0.435 $\pm$ 0.092 \\ \hline
                                               
\multirow{3}{*}{\textsc{\textbf{BERT}}}        & \textbf{N/A}                 & \checkmark & N/A        & 0.844 $\pm$ 0.055 & 0.852 $\pm$ 0.044 & 0.846 $\pm$ 0.055 & 0.840 $\pm$ 0.059 & 0.776 $\pm$ 0.140 & 0.822 $\pm$ 0.060 & 0.789 $\pm$ 0.116 & 0.759 $\pm$ 0.170 \\ \cline{2-12}
                                               & \textsc{\textbf{Node2Vec}}   & \checkmark & \checkmark & 0.814 $\pm$ 0.113 & 0.853 $\pm$ 0.054 & 0.827 $\pm$ 0.099 & 0.806 $\pm$ 0.134 & 0.879 $\pm$ 0.036 & 0.883 $\pm$ 0.036 & 0.885 $\pm$ 0.036 & 0.878 $\pm$ 0.037 \\ \cline{2-12}
                                               & \textsc{\textbf{DeepWalk}}   & \checkmark & \checkmark & 0.825 $\pm$ 0.054 & 0.843 $\pm$ 0.046 & 0.832 $\pm$ 0.049 & 0.823 $\pm$ 0.054 & 0.874 $\pm$ 0.033 & 0.880 $\pm$ 0.027 & 0.877 $\pm$ 0.032 & 0.873 $\pm$ 0.032 \\ \hline
                                               
\multirow{3}{*}{\textsc{\textbf{BERTweet}}}        & \textbf{N/A}                 & \checkmark & N/A        & 0.689 $\pm$ 0.106 & 0.682 $\pm$ 0.208 & 0.687 $\pm$ 0.120 & 0.637 $\pm$ 0.178 & 0.750 $\pm$ 0.080 & 0.806 $\pm$ 0.028 & 0.751 $\pm$ 0.084 & 0.729 $\pm$ 0.121 \\ \cline{2-12}
                                               & \textsc{\textbf{Node2Vec}}   & \checkmark & \checkmark & 0.698 $\pm$ 0.099 & 0.737 $\pm$ 0.089 & 0.680 $\pm$ 0.104 & 0.651 $\pm$ 0.147 & 0.796 $\pm$ 0.033 & 0.812 $\pm$ 0.039 & 0.794 $\pm$ 0.031 & 0.790 $\pm$ 0.031 \\ \cline{2-12}
                                               & \textsc{\textbf{DeepWalk}}   & \checkmark & \checkmark & 0.651 $\pm$ 0.120 & 0.709 $\pm$ 0.172 & 0.666 $\pm$ 0.094 & 0.609 $\pm$ 0.154 & 0.798 $\pm$ 0.047 & 0.820 $\pm$ 0.030 & 0.806 $\pm$ 0.041 & 0.794 $\pm$ 0.051 \\ \hline
                                               
\end{tabular}
}
\end{table*}

For the Twitter16, the best configuration for GETAE is (BERT, \textsc{Node2Vec}, RNN) with 89.6\% Accuracy, 90.1\% Precision, 89.7\% Recall, and 89.5\% F1-Score.
The second best configuration for GETAE is a tie between (BERT, \textsc{Node2Vec}, \textsc{BiRNN}) and (BERT, \textsc{Node2Vec}, \textsc{BiLSTM}) with a difference in metrics smaller than 1\%. 
We observe that for the Twitter16 dataset, the model that employs \textsc{BERTweet} for the word embeddings only performs better when using node embeddings and a \textsc{BiRNN} layer.

\subsection{Node Embeddings Hyperparameter tuning}

For this set of tests, we use the same methodology when training the models using the GETAE as in Subsection~\ref{ssec:class}.

\subsubsection{\textsc{Node2Vec}}

Tables~\ref{tb:n2v15} and~\ref{tb:n2v16} present the detection results when varying Node2Vec parameters $p$ (return hyperparameter), $q$ (in-out hyperparameter), and $d$ (dimensions) for the GETAE architecture.
The return parameter $p$ controls the likelihood of immediately revisiting a node in the walk, while the in-out hyperparameter q allows the search to differentiate between \textit{inward} and \textit{outward} nodes~\cite{Grover2016}.
For each of these configurations of parameters, we use the following configurations for GETAE:
1) (BERT, \textsc{BiLSTM}, \textsc{Node2Vec}) for the Twitter15 dataset, and
2) (BERT, RNN, \textsc{Node2Vec}) for the Twitter16 dataset.

\begin{table}[!htbp]
\centering
\caption{\textsc{Node2Vec} on Twitter15 (BERT + \textsc{BiLSTM})}
\label{tb:n2v15}
\resizebox{\columnwidth}{!}{
\begin{tabular}{|c|c|r|r|r|r|r|}
\hline
\textbf{p} & \textbf{q} & \textbf{d} & \multicolumn{1}{c|}{\textbf{Accuracy}} & \multicolumn{1}{c|}{\textbf{Recall}} & \multicolumn{1}{c|}{\textbf{Precision}} & \multicolumn{1}{c|}{\textbf{F1-Score}} \\ \hline
1.0   & 1.0   & 32  & 0.686 $\pm$ 0.070 & 0.696 $\pm$ 0.056 & 0.691 $\pm$ 0.063 & 0.682 $\pm$ 0.075 \\ \hline
1.0   & 0.5 & 32  & 0.715 $\pm$ 0.022 & 0.724 $\pm$ 0.029 & 0.717 $\pm$ 0.022 & 0.714 $\pm$ 0.022 \\ \hline
0.5 & 1.0   & 32  & 0.701 $\pm$ 0.033 & 0.710 $\pm$ 0.030 & 0.707 $\pm$ 0.032 & 0.700 $\pm$ 0.035 \\ \hline
0.5 & 0.5 & 32  & 0.709 $\pm$ 0.036 & 0.716 $\pm$ 0.039 & 0.711 $\pm$ 0.040 & 0.707 $\pm$ 0.038 \\ \hline
2.0   & 1.0   & 32  & 0.685 $\pm$ 0.056 & 0.704 $\pm$ 0.036 & 0.689 $\pm$ 0.046 & 0.678 $\pm$ 0.066 \\ \hline
1.0   & 2.0   & 32  & 0.713 $\pm$ 0.032 & 0.722 $\pm$ 0.029 & 0.714 $\pm$ 0.030 & 0.712 $\pm$ 0.033 \\ \hline
1.0   & 1.0   & 100 & \textbf{0.827 $\pm$ 0.042} & \textbf{0.831 $\pm$ 0.035} & \textbf{0.827 $\pm$ 0.040} & \textbf{0.825 $\pm$ 0.042} \\ \hline
1.0   & 0.5 & 100 & 0.806 $\pm$ 0.023 & 0.807 $\pm$ 0.024 & 0.804 $\pm$ 0.024 & 0.804 $\pm$ 0.024 \\ \hline
0.5 & 1.0   & 100 & 0.780 $\pm$ 0.042 & 0.790 $\pm$ 0.034 & 0.785 $\pm$ 0.037 & 0.778 $\pm$ 0.044 \\ \hline
0.5 & 0.5 & 100 & 0.823 $\pm$ 0.032 & 0.827 $\pm$ 0.034 & 0.824 $\pm$ 0.034 & 0.821 $\pm$ 0.035 \\ \hline
2.0   & 1.0   & 100 & 0.806 $\pm$ 0.038 & 0.814 $\pm$ 0.037 & 0.809 $\pm$ 0.035 & 0.805 $\pm$ 0.038 \\ \hline
1.0   & 2.0   & 100 & 0.791 $\pm$ 0.051 & 0.813 $\pm$ 0.030 & 0.794 $\pm$ 0.048 & 0.784 $\pm$ 0.058 \\ \hline
\end{tabular}
}
\end{table}

\begin{table}[!htbp]
\centering
\caption{\textsc{Node2Vec} on Twitter16 (BERT + RNN)}
\label{tb:n2v16}
\resizebox{\columnwidth}{!}{
\begin{tabular}{|c|c|r|r|r|r|r|}
\hline
\textbf{p} & \textbf{q} & \multicolumn{1}{c|}{\textbf{d}} & \multicolumn{1}{c|}{\textbf{Accuracy}} & \multicolumn{1}{c|}{\textbf{Recall}} & \multicolumn{1}{c|}{\textbf{Precision}} & \multicolumn{1}{c|}{\textbf{F1-Score}} \\ \hline
1.0   & 1.0   & 32  & 0.826 $\pm$ 0.059 & 0.849 $\pm$ 0.039 & 0.831 $\pm$ 0.053 & 0.822 $\pm$ 0.066 \\ \hline
1.0   & 0.5   & 32  & 0.836 $\pm$ 0.128 & 0.815 $\pm$ 0.198 & 0.842 $\pm$ 0.119 & 0.821 $\pm$ 0.171 \\ \hline
0.5   & 1.0   & 32  & 0.824 $\pm$ 0.091 & 0.796 $\pm$ 0.173 & 0.815 $\pm$ 0.113 & 0.799 $\pm$ 0.151 \\ \hline
0.5   & 0.5   & 32  & 0.861 $\pm$ 0.042 & 0.867 $\pm$ 0.041 & 0.860 $\pm$ 0.040 & 0.856 $\pm$ 0.044 \\ \hline
2.0   & 1.0   & 32  & 0.878 $\pm$ 0.040 & 0.884 $\pm$ 0.034 & 0.873 $\pm$ 0.045 & 0.872 $\pm$ 0.044 \\ \hline
1.0   & 2.0   & 32  & 0.868 $\pm$ 0.042 & 0.874 $\pm$ 0.039 & 0.862 $\pm$ 0.044 & 0.864 $\pm$ 0.044 \\ \hline
1.0   & 1.0   & 100 & \textbf{0.896 $\pm$ 0.025} &\textbf{ 0.901 $\pm$ 0.026} & \textbf{0.897 $\pm$ 0.028} & \textbf{0.895 $\pm$ 0.025} \\ \hline
1.0   & 0.5   & 100 & 0.841 $\pm$ 0.089 & 0.871 $\pm$ 0.040 & 0.841 $\pm$ 0.087 & 0.830 $\pm$ 0.110 \\ \hline
0.5   & 1.0   & 100 & 0.853 $\pm$ 0.071 & 0.868 $\pm$ 0.043 & 0.858 $\pm$ 0.062 & 0.851 $\pm$ 0.074 \\ \hline
0.5   & 0.5   & 100 & 0.822 $\pm$ 0.093 & 0.856 $\pm$ 0.050 & 0.828 $\pm$ 0.078 & 0.813 $\pm$ 0.099 \\ \hline
2.0   & 1.0   & 100 & 0.845 $\pm$ 0.054 & 0.858 $\pm$ 0.044 & 0.846 $\pm$ 0.048 & 0.842 $\pm$ 0.055 \\ \hline
1.0   & 2.0   & 100 & 0.858 $\pm$ 0.049 & 0.864 $\pm$ 0.045 & 0.863 $\pm$ 0.041 & 0.855 $\pm$ 0.049 \\ \hline
\end{tabular}
}

\end{table}

For the Twitter15 dataset (Table~\ref{tb:n2v15}), we observe that the best results are obtained for hyperparameters' values $p = 1$, $q = 1$, and $d = 100$ with $82.7\%$ Accuracy, $83.1\%$ Precision, $82.7\%$ Recall, and $82.5\%$ F1-Score. 
With this GETAE configuration, the node embedding model uses a balanced strategy for random walks, considering both breadth-first and depth-first traversals.
The second best model trained using the GETAE architecture obtained the best results when using $p = 0.5$, $q = 0.5$, and $d = 100$ as hyperparameter values. 
The difference between the best and second best performing models is around 3\%.
This hyperparameters' configuration is not leaning toward breadth-first or depth-first strategies.
Thus, in the case of Twitter15, we notice that having a bias for the random walk does not matter that much, while $d = 100$ tells us that a fine-grained representation is needed for understanding more complex patterns.

For the Twitter16 dataset (Table~\ref{tb:n2v16}), the best model trained using the GETAE architecture uses the following hyperparameters' values $p = 1$, $q = 1$, and $d = 100$. 
This model obtains $89.6\%$ Accuracy, $90.1\%$ Precision, $89.7\%$ Recall, and $89.5\%$ F1-Score.
These hyperparameters offer a balanced random walk strategy.
The second best model is trained using the following hyperparameters' values $p = 2$, $q = 1$, and $d = 32$.
The differences between the best and second best-performing models are between 2-7\% for each metric.
The second best performing model is leaning towards a depth-first sampling strategy due to a higher likeliness of the returning hyperparameter $p$.
Furthermore, the second-best results are obtained when the dimension is small, i.e., $d = 32$.
Thus, we can conclude that a 32-dimensional representation is also not only enough, but better at capturing node information than many other strategies and dimensions.

\subsubsection{\textsc{DeepWalk}}

Tables~\ref{tb:dw15} and~\ref{tb:dw16} present the results obtained by the models trained using the GETAE architecture when using \textsc{DeepWalk} node embeddings~\cite{Perozzi2014}. 
For this set of experiments, we use BERT for embedding the textual data and we vary both the node embedding dimension $d$ as well as the recurrent neural network layer of GETAE's Text Branch.

\begin{table}[!htbp]
\centering
\caption{DeepWalk on Twitter15 + BERT}
\label{tb:dw15}
\resizebox{\columnwidth}{!}{
\begin{tabular}{|r|l|r|r|r|r|}
\hline
\multicolumn{1}{|c|}{\textbf{d}} & \textbf{RNN Layer} & \multicolumn{1}{c|}{\textbf{Accuracy}} & \multicolumn{1}{c|}{\textbf{Recall}} & \multicolumn{1}{c|}{\textbf{Precision}} & \multicolumn{1}{c|}{\textbf{F1-Score}} \\ \hline
32  & \textsc{RNN}    & 0.775 $\pm$ 0.045 & 0.786 $\pm$ 0.036 & 0.771 $\pm$ 0.042 & 0.768 $\pm$ 0.045 \\ \hline
32  & \textsc{BiRNN}  & 0.789 $\pm$ 0.040 & 0.797 $\pm$ 0.032 & 0.789 $\pm$ 0.038 & 0.786 $\pm$ 0.042 \\ \hline
32  & \textsc{GRU}    & 0.744 $\pm$ 0.026 & 0.749 $\pm$ 0.025 & 0.744 $\pm$ 0.027 & 0.743 $\pm$ 0.026 \\ \hline
32  & \textsc{BiGRU}  & 0.765 $\pm$ 0.017 & 0.770 $\pm$ 0.014 & 0.767 $\pm$ 0.019 & 0.764 $\pm$ 0.018 \\ \hline
32  & \textsc{LSTM}   & 0.758 $\pm$ 0.043 & 0.787 $\pm$ 0.033 & 0.758 $\pm$ 0.034 & 0.752 $\pm$ 0.043 \\ \hline
32  & \textsc{BiLSTM} & 0.805 $\pm$ 0.037 & \textbf{0.819 $\pm$ 0.034} & 0.808 $\pm$ 0.038 & 0.801 $\pm$ 0.039 \\ \hline
100 & \textsc{RNN}    & 0.736 $\pm$ 0.063 & 0.773 $\pm$ 0.044 & 0.738 $\pm$ 0.060 & 0.725 $\pm$ 0.071 \\ \hline
100 & \textsc{BiRNN}  & 0.795 $\pm$ 0.031 & 0.802 $\pm$ 0.032 & 0.796 $\pm$ 0.030 & 0.791 $\pm$ 0.031 \\ \hline
100 & \textsc{GRU}    & 0.751 $\pm$ 0.036 & 0.751 $\pm$ 0.036 & 0.750 $\pm$ 0.036 & 0.749 $\pm$ 0.035 \\ \hline
100 & \textsc{BiGRU}  & 0.776 $\pm$ 0.015 & 0.776 $\pm$ 0.017 & 0.775 $\pm$ 0.014 & 0.773 $\pm$ 0.016 \\ \hline
100 & \textsc{LSTM}   & 0.764 $\pm$ 0.050 & 0.782 $\pm$ 0.035 & 0.760 $\pm$ 0.050 & 0.757 $\pm$ 0.059 \\ \hline
100 & \textsc{BiLSTM} & \textbf{0.814 $\pm$ 0.043} & 0.818$\pm$0.041 & \textbf{0.815$\pm$0.042} & \textbf{0.812 $\pm$ 0.044} \\ \hline
\end{tabular}
}
\end{table}

\begin{table}[!htbp]
\centering
\caption{DeepWalk on Twitter16 + BERT}
\label{tb:dw16}
\resizebox{\columnwidth}{!}{
\begin{tabular}{|c|l|r|r|r|r|}
\hline
\multicolumn{1}{|c|}{d} & \textbf{RNN Layer} & \multicolumn{1}{|c|}{\textbf{Accuracy}} & \multicolumn{1}{|c|}{\textbf{Recall}} & \multicolumn{1}{|c|}{\textbf{Precision}} & \multicolumn{1}{|c|}{\textbf{F1-Score}} \\ \hline
32  & \textsc{RNN}    & 0.844 $\pm$ 0.098 & 0.870 $\pm$ 0.068 & 0.847 $\pm$ 0.090 & 0.838 $\pm$ 0.108 \\ \hline
32  & \textsc{BiRNN}  & 0.843 $\pm$ 0.095 & 0.860 $\pm$ 0.064 & 0.845 $\pm$ 0.083 & 0.838 $\pm$ 0.102 \\ \hline
32  & \textsc{GRU}    & 0.853 $\pm$ 0.039 & 0.861 $\pm$ 0.036 & 0.852 $\pm$ 0.039 & 0.852 $\pm$ 0.040 \\ \hline
32  & \textsc{BiGRU}  & 0.846 $\pm$ 0.026 & 0.853 $\pm$ 0.021 & 0.849 $\pm$ 0.024 & 0.846 $\pm$ 0.026 \\ \hline
32  & \textsc{LSTM}   & 0.860 $\pm$ 0.053 & 0.870 $\pm$ 0.044 & 0.859 $\pm$ 0.050 & 0.856 $\pm$ 0.056 \\ \hline
32  & \textsc{BiLSTM} & \textbf{0.881 $\pm$ 0.025} & 0.880 $\pm$ 0.023 & \textbf{0.879 $\pm$ 0.025} & \textbf{0.880 $\pm$ 0.026} \\ \hline
100 & \textsc{RNN}    & 0.801 $\pm$ 0.072 & 0.836 $\pm$ 0.037 & 0.800 $\pm$ 0.079 & 0.789 $\pm$ 0.089 \\ \hline
100 & \textsc{BiRNN}  & 0.851 $\pm$ 0.040 & 0.856 $\pm$ 0.036 & 0.856 $\pm$ 0.032 & 0.850 $\pm$ 0.038 \\ \hline
100 & \textsc{GRU}    & 0.845 $\pm$ 0.031 & 0.849 $\pm$ 0.025 & 0.847 $\pm$ 0.030 & 0.843 $\pm$ 0.032 \\ \hline
100 & \textsc{BiGRU}  & 0.857 $\pm$ 0.027 & 0.866 $\pm$ 0.024 & 0.862 $\pm$ 0.031 & 0.856 $\pm$ 0.027 \\ \hline
100 & \textsc{LSTM}   & 0.825 $\pm$ 0.054 & 0.843 $\pm$ 0.046 & 0.832 $\pm$ 0.049 & 0.823 $\pm$ 0.054 \\ \hline
100 & \textsc{BiLSTM} & 0.874 $\pm$ 0.033 & \textbf{0.880 $\pm$ 0.027} & 0.877 $\pm$ 0.032 & 0.873 $\pm$ 0.032 \\ \hline
\end{tabular}
}
\end{table}

\begin{table*}[!htbp]
\caption{Fake News Detection Comparison with State-of-the-Art Models}
\label{tb:bestm}
\centering
\begin{tabular}{|l|c|c|c|c|c|c|c|c|}
\hline
\multirow{2}{*}{\textbf{Model}} & \multicolumn{4}{c|}{\textbf{Twitter15}} & \multicolumn{4}{c|}{\textbf{Twitter16}} \\ \cline{2-9}
                                                    & \textbf{Accuracy} & \textbf{Precision} & \textbf{Recall} & \textbf{F1-Score} & \textbf{Accuracy} & \textbf{Precision} & \textbf{Recall} & \textbf{F1-Score} \\ \hline
\multicolumn{1}{|l|}{DTC~\cite{Castillo2011}}       & 0.4949            & 0.4963             & 0.4806          & 0.4948            & 0.5612            & 0.5753             & 0.5369          & 0.5616            \\ \hline
\multicolumn{1}{|l|}{SVM-TS~\cite{Ma2016detecting}} & 0.5195            & 0.5195             & 0.5186          & 0.5190            & 0.6932            & 0.6928             & 0.6910          & 0.6915            \\ \hline
\multicolumn{1}{|l|}{mGRU~\cite{Ma2016detecting}}   & 0.5547            & 0.5145             & 0.5148          & 0.5104            & 0.6612            & 0.5603             & 0.5618          & 0.5563            \\ \hline
\multicolumn{1}{|l|}{RFC~\cite{Kwon2017}}           & 0.5385            & 0.5718             & 0.5302          & 0.4642            & 0.6620            & 0.7315             & 0.6587          & 0.6275            \\ \hline
\multicolumn{1}{|l|}{tCNN~\cite{Yang2018}}          & 0.5881            & 0.5199             & 0.5206          & 0.5140            & 0.7374            & 0.6248             & 0.6262          & 0.6200            \\ \hline
\multicolumn{1}{|l|}{CRNN~\cite{Liu2018}}           & 0.5919            & 0.5296             & 0.5305          & 0.5249            & 0.7576            & 0.6419             & 0.6433          & 0.6367            \\ \hline
\multicolumn{1}{|l|}{CSI~\cite{Ruchansky2017}}      & 0.6987            & 0.6991             & 0.6867          & 0.7174            & 0.6612            & 0.6321             & 0.6309          & 0.6304            \\ \hline
\multicolumn{1}{|l|}{dEFEND~\cite{Shu2019}}         & 0.7383            & 0.6584             & 0.6611          & 0.6541            & 0.7016            & 0.6365             & 0.6384          & 0.6311            \\ \hline
\multicolumn{1}{|l|}{DANES~\cite{Truica2023danes}}  & 0.7790            & 0.7856             & 0.7800          & 0.7827            & 0.7795            & 0.7854             & 0.7795          & 0.7824            \\ \hline
\multicolumn{1}{|l|}{GCAN-G~\cite{Lu2020GCAN}}      & 0.8636            & 0.7959             & 0.7990          & 0.7938            & 0.7939            & 0.6785             & 0.6802          & 0.6754            \\ \hline
\multicolumn{1}{|l|}{GCAN~\cite{Lu2020GCAN}}        & \textbf{0.8767}   & 0.8257             & 0.8295          & 0.8250            & \textbf{0.9084}   & 0.7594             & 0.7632          & 0.7593            \\ \hline
\multicolumn{1}{|l|}{\textbf{GETAE}}                & 0.825             & \textbf{0.827}     & \textbf{0.831}  & \textbf{0.827}    & 0.895             & \textbf{0.897}     & \textbf{0.901}  & \textbf{0.896}    \\ \hline

\end{tabular}
\end{table*}

For the Twitter15 dataset (Table~\ref{tb:dw15}), the best results are obtained when using $d = 100$ for the node embedding and the \textsc{BiLSTM} layer for the Text Branch.
The model trained using this configuration obtains 81.4\% Accuracy, 81.8\% Precision, 81.5\% Recall, and 81.2\% F1-Score on the test set.
The second best model obtains the best results when the hyperparameters configuration is $d = 32$ and a \textsc{BiLSTM}
The differences between the best and second best-performing models are between 1-2\% for each metric.
We observe that the models trained using the GETAE architecture that employs $d = 100$ obtain better results than when using $d=32$.
We conclude that having a higher number of dimensions yields better results when training models with the GETAE architecture for this dataset.
However, we observe that choosing to use a \textsc{BiLSTM} layer for GETAE's Text Branch has a greater influence than an increase in dimensions of the node embedding for GETAE's Propagation Branch.

For the Twitter16 dataset (Table~\ref{tb:dw16}), we observe that smaller node embedding dimensions are better for training models with the GETAE architecture.
The best model uses the \textsc{BiLSTM} for GETAE's Text Branch and a 32-dimensional node embedding for GETAE's Propagation Branch. 
This model obtains 88.1\% Accuracy, 88\% Precision, 87.9\% Recall, and 88\% F1-Score on the test set.
In comparison, the results obtained by this model are around 1\% better than when using $d=100$ for the node embedding.
We can conclude that for this dataset, the difference between using a lower or higher dimension representation of the node embedding changes very little the models' overall performance. 

\subsection{Comparison with State-of-the-Art Models}

To better determine the performance of GETAE, we compare the results obtained by our model with state-of-the-art models trained on the Twitter15 and Twitter16 datasets.
In our comparison, we employ the following models also used by~\cite{Lu2020GCAN} in their comparison:
\begin{itemize}
    \item\textbf{DTC}~\cite{Castillo2011} uses Decision Trees from user and tweet data.
    \item\textbf{SVM-TS}~\cite{Ma2016detecting} uses a Support Vector Machine with data from the source tweet and the retweets sequence.
    \item\textbf{mGUR}~\cite{Ma2016detecting} is a modified GRU neural network using the source tweet and the patterns in retweets.
    \item\textbf{RFC}~\cite{Kwon2017} is a Random Forest-based classifier that learns on the source tweet and user retweets.
    \item\textbf{tCNN}~\cite{Yang2018} is a modified convolutional neural network that learns local patterns of user profile sequence and tweet data.
    \item\textbf{CRNN}~\cite{Liu2018} is an RNN and CNN approach using the source tweet data.
    \item\textbf{CSI}~\cite{Ruchansky2017} uses the tweets and a scoring mechanism of the users that retweet them.
    \item\textbf{dEFEND}~\cite{Shu2019} is a co-attention-based model that correlates the source tweet and the user profile.
    \item\textbf{DANES}~\cite{Truica2023danes} is a model that uses a new node embedding from social and textual context data. 
    \item\textbf{GCAN} and \textbf{GCAN-G}~\cite{Lu2020GCAN}. GCAN uses user features and combines them with recurrent and convolutional networks for retweet propagation. GCAN-G learns the same features but without the convolutional network.
\end{itemize}

Table~\ref{tb:bestm} presents the results of our comparison. 
On Twitter15, GETAE outperforms the other model obtaining 82.7\% F1-Score, 83.1\% Recall, 82.7\% Precision, while on Twitter16 obtaining 89.6\% F1-Score, 90.1\% Recall and 89.7\% Precision.
The best accuracies are obtained by GCAN, while the second best results come from GCAN, DANES, and GCAN-G.
Except for DANES, both GCAN and GCAN-G are large deep neural networks.

For testing our best node embedding techniques, we compared GETAE with two state-of-the-art models:
\begin{itemize}
    \item\textbf{HiMaP}~\cite{Mishra2020} learns indirect relationships between users with graph embeddings.
    \item\textbf{SSGE}~\cite{Vu2021rumor} uses graph embeddings to learn propagation representations.
\end{itemize}

Table~\ref{tb:bestne} presents the comparison between the two state-of-the-art models and GETAE. 
On the Twitter15 dataset, GETAE obtained the best results when using \textsc{Node2Vec}.
We observe that our model is slightly surpassed by HiMaP with a small difference in Accuracy of only 1.9\%.
On the Twitter16 dataset, GETAE obtains the overall best performance with an Accuracy of 89.6\%, outperforming the other state-of-the-art models used in this comparison.

\begin{table}[!htbp]
\caption{Comparison with State-of-the-Art w.r.t. Node Embeddings}
\label{tb:bestne}
\centering
\begin{tabular}{|l|c|c|c|}
\hline
\textbf{Model} & \multicolumn{1}{|c|}{\textbf{\begin{tabular}[c]{@{}c@{}}Node\\ Embedding\end{tabular}}} & \multicolumn{1}{c|}{\textbf{\begin{tabular}[c]{@{}c@{}}Twitter15\\ Accuracy\end{tabular}}} & \multicolumn{1}{c|}{\textbf{\begin{tabular}[c]{@{}c@{}}Twitter16\\ Accuracy\end{tabular}}} \\ \hline
\multirow{2}{*}{HiMaP~\cite{Mishra2020}}     & Node2Vec & \textbf{0.846} & 0.867                          \\ \cline{2-4} 
                           & DeepWalk & 0.825          & 0.850                          \\ \hline
\multirow{2}{*}{SSGE~\cite{Vu2021rumor}}      & Node2Vec & 0.600          & 0.600                          \\ \cline{2-4} 
                           & DeepWalk & 0.500          & 0.500                          \\ \hline
\multirow{2}{*}{\textbf{GETAE}} & Node2Vec & 0.827          & \textbf{0.896}                 \\ \cline{2-4} 
                           & DeepWalk & 0.814          & 0.881                          \\ \hline
\end{tabular}
\end{table}

\section{Discussion and Limitations}~\label{sec:discussion}

GETAE is a promising ensemble architecture for training fake news detection models that consider the information propagation of content.
By considering how harmful nodes infect the network, the models trained outperform existing state-of-the-art models. 

On Twitter15, in terms of Accuracy and F1-Score, the model trained using \textsc{BiLSTM} for the \textsc{[Bi]RNN} layer, \textsc{BERT} as the word embedding model, and \textsc{Node2Vec} as the node embedding outperforms the other models.
Using this configuration for the GETAE architecture, the trained model obtains an Accuracy score of $82.7\%$ and an F1-Score of $82.5\%$.
When configuring the GETAE architecture to use during training \textsc{BiRNN} for the \textsc{[Bi]RNN} layer, \textsc{BERTweet} as the word embedding model, and \textsc{DeepWalk} as the node embedding, we obtain a Precision score of $83.9\%$ and an F1-Score of $82.7\%$.
During hyperparameter tuning, when using the \textsc{BiLSTM} for the \textsc{[Bi]RNN} layer and \textsc{BERT} as the word embedding model, we observe that the best hyperparameters for \textsc{Node2Vec} are $p = 1, q = 1, dim = 100$ followed by $p = 0.5, q = 0.5, dim = 100$ with a lower mean accuracy by $\sim 4\%$, but a higher standard deviation by $\sim 0.1\%$. \textsc{Deepwalk} also provided the best overall results when using the BERT as the word embeddings and \textsc{BiLSTM} for the \textsc{[Bi]RNN} layer.
The best GETAE configuration that uses \textsc{Word2Vec} as the word embedding mode employs \textsc{BiRNN} for the \textsc{[Bi]RNN} layer and no node embedding.
For this dataset, we can conclude that the \textit{Propagation-Enhanced Content Embedding} performs better when using contextual word embeddings, such as BERT and BERTweet, as well as a bidirectional approach for the \textsc{[Bi]RNN} layer.

For the Twitter16 dataset, the model trained using \textsc{RNN} for the \textsc{[Bi]RNN} layer, \textsc{BERT} as the word embedding model, and \textsc{Node2Vec} as the node embedding outperforms the other models.
With this configuration, the model obtains an Accuracy score of $89.6\%$, a Precision score of $90.1\%$, a Recall score of $89.7\%$, and an F1-Score of $89.5\%$.
When configuring the GETAE architecture to use during training \textsc{BiRNN} for the \textsc{[Bi]RNN} layer, \textsc{BERTweet} as the word embedding model, and \textsc{DeepWalk} as the node embedding, we obtain an Accuracy score of $82.9\%$, Precision score of $84.2\%$, Recall score of $83.1\%$ and an F1-Score of $82.6\%$.
During hyperparameter tuning, when using the \textsc{RNN} for the \textsc{[Bi]RNN} layer and \textsc{BERT} as the word embedding model, we observe that the best hyperparameters for \textsc{Node2Vec} are $p = 1, q = 1, dim = 100$ followed by $p = 2, q = 1.0, dim = 32$ with a lower mean accuracy by $\sim 3\%$.
\textsc{Deepwalk} also provided the best overall results when using the BERT as the word embeddings and \textsc{BiLSTM} for the \textsc{[Bi]RNN} layer for both small and large vector dimensions.
The best GETAE configuration that uses \textsc{Word2Vec} as the word embedding mode employs \textsc{BiRNN} for the \textsc{[Bi]RNN} layer and \textsc{Node2Vec} as node embedding.

In comparison with other state-of-the-art models on Fake News Detection or Rumor Detection, our best models outperform GCAN~\cite{Lu2020GCAN} in terms of Precision, Recall, and F1-Score, but it is outperformed in terms of Accuracy.
As we used Stratified K-Fold validation and conducted 10 experiments for each measure while randomizing the train-test-validation split each time, we observed that the standard deviation was small enough to conclude that our models generalize well on the two datasets. 
During hyperparameter tuning, we obtain better results on high-dimensional node embeddings. Thus, we can conclude that more detailed representations are needed for capturing the topological information of the graph as well as better representing the information diffusion in order to create better \textit{Propagation-Enhanced Content Embeddings}.
Furthermore, for the \textsc{Node2Vec}, the best return parameters and in-out parameters show that a balanced random walk strategy is preferred, with depth-first search in second place. 
Based on these observations, we can conclude that:
\begin{itemize}
    \item The choice of models for word embeddings, node embeddings, and recurrent layers needs to be done after proper testing, in order to get a better fit to the data at hand --- answering \textit{Q1}.
    \item The Text Content Embedding manages to better encode multiple complex textual features such as lexical (e.g., character and word level features) and syntactic (e.g., sentence-level features) to improve prediction --- answering \textit{Q2}.
    \item The Propagation Embedding is specifically designed to encode the information diffusion to determine the spread of information from a node to its followers --- answering \textit{Q3}.
    \item The Propagation-Enhanced Content Embedding manages to better encapsulate textual content and context and the propagation information to improve fake news detection --- answering \textit{Q4}.
    \item Our data-driven best models trained with GETAE manage to outperform existing state-of-the-art models trained on the same datasets --- answering \textit{Q5}.
\end{itemize}

We conclude the discussion by addressing one final question: \textit{Can a propagation-enhanced deep neural network ensemble architecture 
provide a definitive solution to the challenge of Fake News Detection?}
The answer is a qualified no.
While the proposed architecture offers potential improvements, no single solution can fully address the multifaceted nature of fake news.
A mixture of models and techniques, e.g., Mixture of Experts (MoE), is necessary to improve the accuracy of predicting fake news.
In such ecosystems, the proposed GETAE architecture demonstrates promising accuracy when incorporating information propagation and textual context.
Numerous challenges remain and need to be addressed to help users make informed decisions while browsing news feeds on social networks.

The primary challenge we identified through our review of the current literature includes:
\begin{itemize}
\item[(\textit{1})] Stable datasets that preserve all metadata information, rather than relying on unstable URLs and unique IDs.
\item[(\textit{2})] Multi-modal datasets on which we can train richer feature representations.
\item[(\textit{3})] Human verified dataset that can enhance feature selection~\cite{Parikh2018media}.
\end{itemize}
As a final thought, a significant limitation in this research area is the absence of large, annotated datasets that capture social media content and underlying network characteristics, such as post reactions, diffusion, etc. 
Such datasets would enable us to train and fine-tune our models for enhanced performance.

\section{Conclusions}~\label{sec:conclusions}

In this paper, we propose GETAE, \underline{G}raph Information \underline{E}nhanced Deep Neural Ne\underline{t}work Ensemble \underline{A}rchitectur\underline{E} for Fake News Detection, a novel ensemble architecture that uses textual content together with the social interactions to improve fake news detection.
We propose 3 types of embeddings, 
(1) a Text Content Embedding that manages to better encode lexical and syntactic textual features, 
(2) a Propagation Embedding specifically designed to encode the information diffusion, and
(3) a Propagation-Enhanced Content Embedding that encapsulates both textual and propagation information.
Using cross-validation, ablation testing, and hyperparameter tuning, we train multiple models on two real-world datasets, i.e., Twitter15 and Twitter16.
We observe that the models trained using GETAE manage to outperform the current state-of-the-art, showing the importance of integrating network and diffusion information into the training step to improve inference.

BERT and \textsc{BERTweet} Transformer embeddings paired with \textsc{Node2Vec} or \textsc{DeepWalk} node embeddings and \textsc{BiLSTM} or RNN recurrent layers gained a meaningful and accurate representation of both tweet content and network topology of the source tweets.
We conclude that for \textsc{Node2Vec} there is no simple strategy for biased random walks, our results despite having good accuracy, have no consistency leaning towards breadth-first or depth-first strategies.
Our best recurrent layers in terms of accuracy were \textsc{BiLSTM} for Twitter15 and RNN for Twitter16 paired with BERT Transformer embeddings for text vectorization. 

In the future, we plan to incorporate additional context-based features, such as the number of likes, retweets, and comments, into this model.
We aim to add new convolutional layers to the GETAE architecture in order to better understand the topological patterns of the source nodes that spread fake news.
Moreover, we plan to experiment on larger datasets and other social networks where we can create domain-specific lexicon techniques~\cite{Truica2023ATE} for misinformation and propaganda detection and mitigation.
Finally, based on our previous approaches, we aim to create a Mixture of Experts (MoE) based solution for fake news detection. 

\section*{Acknowledgments}
The research presented in this paper was supported in part by 
(1) a grant from the National Program for Research of the National Association of Technical Universities (GNAC ARUT 2023) through the project ``DEPLATFORM: Intelligent interactive system for detecting the veracity of news published on social platforms'' (Contract no. 63/10.10.2023),  
(2) the German Academic Exchange Service (DAAD) through the project ``iTracing: Automatic Misinformation Fact-Checking'' (DAAD grant no. 91809005), 
(3) The Academy of Romanian Scientists through the funding of project ``SCAN-NEWS: Smart system for deteCting And mitigatiNg misinformation and fake news in social media'' (AOŞR-TEAMS-III),
...

\bibliographystyle{elsarticle-num-names} 
\bibliography{main}

\end{document}